\documentclass{article}
\usepackage[a4paper, margin=1in, footskip=0.5in, tmargin=0.9in, bmargin=1in]{geometry}
\usepackage{natbib}
\usepackage[table]{xcolor}
\definecolor{mediumblue}{RGB}{100, 149, 237}
\usepackage{colortbl}
\usepackage{titletoc}
\usepackage{latexsym}
\usepackage{amsmath}
\usepackage{amssymb}
\usepackage{amsthm}
\usepackage[backref=page]{hyperref}
\usepackage{newtxtt}
\usepackage{subcaption}

\hypersetup{
    colorlinks=true,
    linkcolor=[rgb]{1,0.0,0.5},
    filecolor=magenta,      
    urlcolor=[rgb]{0.3,0.75,0.7},
    citecolor=[rgb]{0.1,0.3,0.88}
    }

\usepackage{epsfig}
\usepackage[shortlabels]{enumitem}
\everymath=\expandafter{\the\everymath\displaystyle}
\DeclareMathSizes{24}{24}{24}{24}

\usepackage{color-edits}
\addauthor[Vatsal]{vs}{magenta}
\addauthor[Kameron]{ks}{pink}
\addauthor[Bhavya]{b}{red}

\usepackage{amsthm}
\usepackage{bbm,dsfont}
\usepackage{booktabs,siunitx}
\usepackage{multirow}
\usepackage{isomath}
\usepackage{upgreek}
\usepackage{graphicx}
\usepackage{wrapfig}
\usepackage{cite}
\usepackage{nicefrac}       
\usepackage{microtype}      
\usepackage{xcolor}         

\usepackage[utf8]{inputenc} 
\usepackage[T1]{fontenc}    
\usepackage{url}            
\usepackage{booktabs}       
\usepackage{amsmath,amsfonts,amsthm,amssymb,amsopn,bm,amstext}       
\usepackage{nicefrac}       
\usepackage{microtype}      
\usepackage{xcolor} 
\usepackage{comment}
\usepackage{graphics}
\newcommand{\vct}[1]{\boldsymbol{#1}} 

\newcommand{\field}[1]{\mathbb{#1}}
\newcommand{\R}{\field{R}} 
\newcommand{\E}{\field{E}} 

\newcommand{\name}{CMID }

\DeclareMathOperator*{\argmax}{arg\,max}
\DeclareMathOperator*{\argmin}{arg\,min}

\newtheorem{theorem}{Theorem}

\newtheorem{assumption}{Assumption}

\newtheorem{propo}{Proposition}
\newtheorem{definition}{Definition}

\newtheorem*{remark*}{Remark}

\newcommand{\w}{{\vct{w}}}

\newcommand{\calE}{{\mathcal{E}}}
\newcommand{\calC}{{\mathcal{C}}}

\newcommand{\calR}{{\mathcal{R}}}
\newcommand{\calM}{{\mathcal{M}}}
\newcommand{\calI}{{\mathcal{I}}}

\newcommand{\calD}{{\mathcal{D}}}
\newcommand{\fta}{X_1}
\newcommand{\ftb}{X_2}
\newcommand{\ftc}{X_3}
\newcommand{\wa}{w_1}
\newcommand{\wb}{w_2}
\newcommand{\wc}{w_3}
\newcommand{\ma}{\mu_1}
\newcommand{\mb}{\mu_2}
\newcommand{\mc}{\mu_3}
\newcommand{\eps}{c}
\newcommand{\sia}{\sigma_1}
\newcommand{\sib}{\sigma_2}
\newcommand{\sic}{\sigma_3}

\newcommand{\mdl}{M}
\newcommand{\mtil}{M}
\newcommand{\ytil}{Y}
\newcommand{\lm}{\lambda}
\newcommand{\lr}{\text{LR}}
\newcommand{\bs}{\text{BS}}

\newcommand{\sca}{S}
\newcommand{\ts}[1]{{\fontfamily{txtt}\fontseries{lc}\fontshape{n}\selectfont #1}}

\newcommand{\tsc}[1]{{\fontfamily{pcr}\selectfont #1}}

\newcommand{\ind}{\mathds{1}}

\usepackage{xparse}
\NewDocumentCommand{\colorlinkmailto}{m m m}{%
    \href{mailto:#1}{\textcolor{#2}{#3}}%
}

\title{Mitigating Simplicity Bias in Deep Learning for\\ Improved OOD Generalization and Robustness}

\makeatletter
\def\@fnsymbol#1{\ensuremath{\ifcase#1\or \dagger\or \ddagger\or
   \mathsection\or \mathparagraph\or \|\or **\or \dagger\dagger
   \or \ddagger\ddagger \else\@ctrerr\fi}}
    \makeatother

\author{
Bhavya Vasudeva \quad \quad Kameron Shahabi \quad \quad Vatsal Sharan\\
{\fontsize{9}{11}\selectfont\texttt{\{}\colorlinkmailto{bvasudev@usc.edu}{black}{\texttt{bvasudev}}\texttt{,}   \colorlinkmailto{kyshahab@usc.edu}{black}{\texttt{kyshahab}}\texttt{,}   \colorlinkmailto{vsharan@usc.edu}{black}{\texttt{vsharan}}\texttt{\}@usc.edu}}\\
University of Southern California
}

\date{}
\begin{document}

\maketitle

\begin{abstract}
  Neural networks (NNs) are known to exhibit \emph{simplicity bias} where they tend to prefer learning `simple' features over more `complex' ones, even when the latter may be more informative. Simplicity bias can lead to the model making biased predictions which have poor out-of-distribution (OOD) generalization. To address this, we propose a framework that encourages the model to use a more diverse set of features to make predictions. We first train a simple model, and then regularize the conditional mutual information with respect to it to obtain the final model. We demonstrate the effectiveness of this framework in various problem settings and real-world applications, showing that it effectively addresses simplicity bias and leads to more features being used, enhances OOD generalization, and improves subgroup robustness and fairness. We complement these results with theoretical analyses of the effect of the regularization and its OOD generalization properties.
\end{abstract}

\section{Introduction}\label{sec:intro}
Motivated by considerations of understanding generalization in deep learning, there has been a series of interesting studies \citep{zhang2017understanding,lottery,NakkiranDescent} on understanding function classes favored by current techniques for training large neural networks. An emerging hypothesis is that deep learning techniques prefer to learn simple functions over the data. While this inductive bias has benefits in terms of preventing overfitting and improving (in-distribution) generalization in several cases, it is not effective in all scenarios. Specifically, it has been found that in the presence of multiple predictive features of varying complexity, neural networks tend to be overly reliant on simpler features while ignoring more complex features that may be equally or more informative of the target \citep{pitfalls,inc-comp,sb1layer}. This phenomenon has been termed \emph{simplicity bias} and has several undesirable implications for robustness and out-of-distribution (OOD) generalization.

\begin{figure}[th!]
    \centering
    \includegraphics[height=180pt]{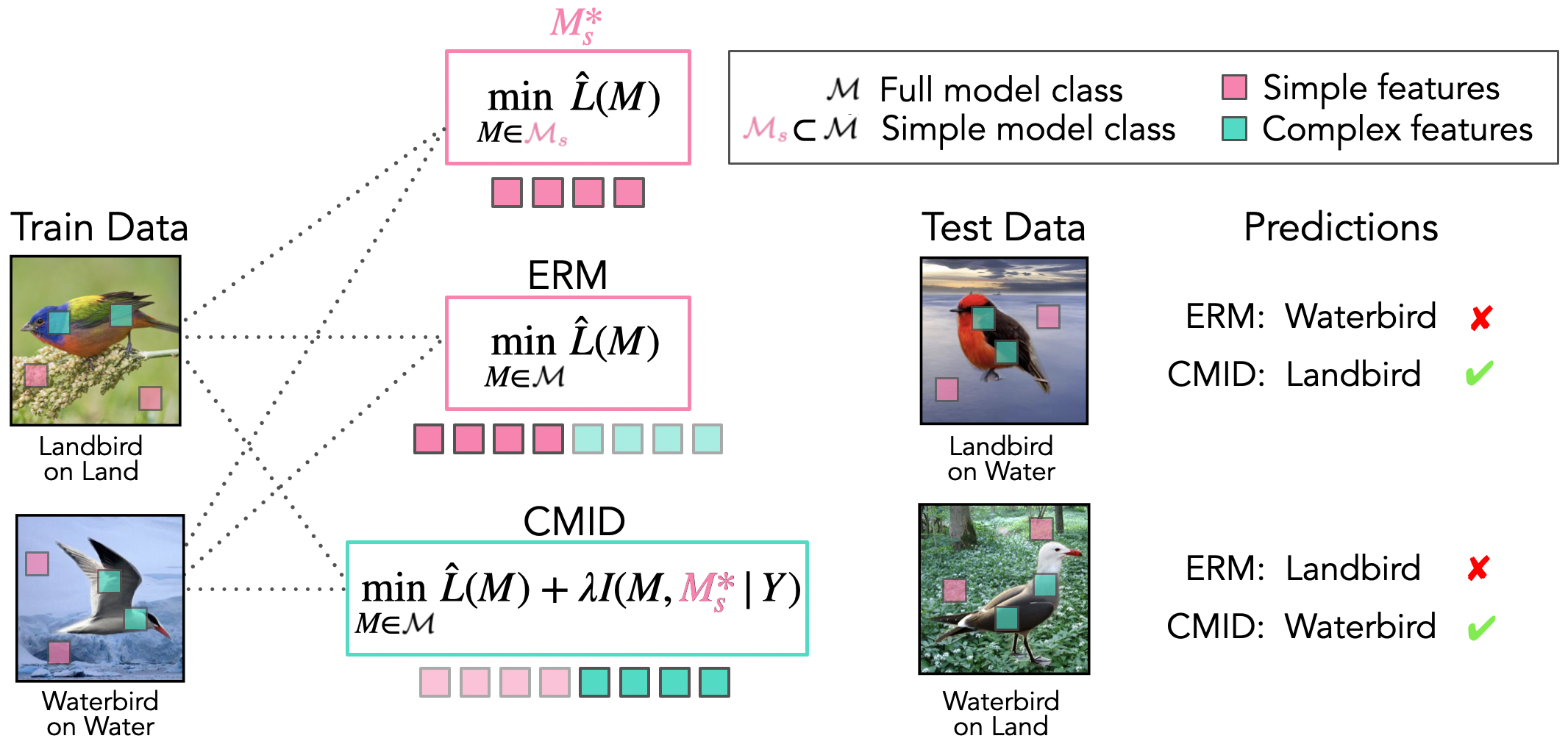}  
    \caption{Summary of our approach. Models trained with ERM tend to focus on simple features (such as background) that do not generalize, whereas encouraging conditional independence with respect to a simple model increases reliance on complex features (such as shape) that generalize.}
    \label{fig:introfig}
\end{figure}

As an illustrative example, consider the {Waterbirds} dataset \citep{gdro}.
The objective here is to predict a bird's type (landbird vs. waterbird) based on its image (see Fig. \ref{fig:introfig} for an example). Note that features such as the background (land vs water background) are easier to learn, and can have significant correlation with the bird's type (since most images of landbirds are on a land background, and vice versa). Although features such as the shape of the bird are more predictive of its type, these tend to be more complex. Simplicity bias can cause the model to be highly dependent on simpler features which are also predictive, such as the background in this case. A model which puts high emphasis on the background for solving this task is not desirable, since its performance may not transfer across different environments. A similar story arises in many different tasks---Table \ref{tab1} summarizes various datasets, where the target or task-relevant features (also known as invariant features in the literature), are more complex than surrogate features that are superficially correlated with the label (also known as spurious features). In such cases, simplicity bias causes NNs to heavily rely on these surrogate features for predictions.
\begin{table}[h!]
    \centering
    \resizebox{0.9\columnwidth}{!}{%
\begin{tabular}{ccc}
\hline
   Dataset & Task-relevant/invariant feature & Surrogate/spurious feature\\
   \hline
   Waterbirds 
   & Bird type (waterbird/landbird) & Background (water/land) \\
   CelebA 
   & Hair colour (blonde/other) & Gender (female/male) \\
   MultiNLI 
   & Reasoning (entailment/neutral/contradiction) & Negation words (`not' etc.)\\
   CivilComments-WILDS 
   & Sentiment (toxic/non-toxic) & Demographic attributes (race, gender, religion) \\
   Colored-MNIST 
   & Digit ($<5$ or $\geq 5$) & Color (red/green)\\
   Camelyon17-WILDS 
   & Diagnoses (tumor/no tumor) & Hospital \\
   Adult-Confounded 
   & Income ($<\$50k$ or $\geq \$50k$) & Demographic attributes (race, gender) \\
   \hline
\end{tabular}
}\caption{Summary of the datasets we consider. Spurious features seem \emph{simpler} than invariant features.
}
\label{tab1}
\end{table}

Several methods \citep{irm,eiil,blood,gao-ood} have been proposed to address this problem of OOD generalization. However, most of them require some knowledge about the spurious feature. This is because some knowledge of the different environments of interests or the underlying causal graph is in general necessary to decide whether a feature is spurious or invariant (due to the No Free Lunch Theorem, see Section~\ref{sec:discussion} for more discussion). To sidestep this, \emph{in this work we take the viewpoint that features that are usually regarded as being spurious for the task are often simple and quite predictive} (as suggested by Table \ref{tab1}). If this hypothesis is true, alleviating simplicity bias can lead to better robustness and OOD generalization.

With improved robustness and OOD generalization as major end goals in mind, we develop a new framework to mitigate simplicity bias and encourage the model to utilize a more diverse set of features for its predictions. Our high level goal is to ensure that the predictions of the trained model $M$ have minimal conditional mutual information $I(M;S|Y)$\footnote{With slight abuse of notation, $M$ denotes both the model and the random variable associated with its predictions.}
with any simple, predictive feature $S$, conditioned on the label $Y$. To achieve this, we first train a simple model $M_s^*$ on the task, with the idea that this model captures the information in simple, predictive features. Subsequently, to train the final model $M$ we add its conditional mutual information $I(M;M_s^*|Y)$ with the model $M_s^*$ conditioned on the label $Y$ as a regularizer to the usual empirical risk minimization (ERM) objective. With this regularization term, we incentivize the model to leverage additional, task-relevant features which may be more complex. We refer to our approach as \emph{Conditional Mutual Information Debiasing or CMID}
(see Fig. \ref{fig:introfig}). 

\emph{We demonstrate that our framework effectively mitigates simplicity bias, and achieves improved OOD generalization, sub-group robustness and fairness across a number of datasets}. We show that it leads to models which use more diverse features on certain tasks which have been previously proposed to measure simplicity bias. The method achieves improved sub-group robustness and OOD generalization on several benchmark tasks, including in the presence of multiple spurious features. It also leads to improved predictions from the perspective of fairness since the predictions are less dependent on protected attributes such as race or gender. In addition, \emph{we prove theoretical results which provide insight and certain guarantees for our approach}. We analyze a Gaussian mixture model setup to understand the effect of our regularization and its ability to reduce dependence on spurious features. We also connect our approach to the causal learning literature by deriving guarantees on OOD generalization in a causal learning framework. We summarize our contributions as follows:

\begin{itemize}

\item We empirically evaluate and validate the hypothesis that spurious features are simpler than invariant features, with respect to representative datasets from Table \ref{tab1} (Section \ref{sec:define_simple}). Based on our findings, we propose a novel framework (CMID) to mitigate simplicity bias and encourage the model to utilize a diverse set of features for its predictions (Section \ref{sec:cmid}).

\item Empirically, we demonstrate that our framework effectively mitigates simplicity bias, and achieves improved OOD generalization, sub-group robustness and fairness across 10 benchmark datasets (Section \ref{sec:expts}). These include different modalities such as image, text, and tabular data and application domains such as healthcare and fairness. While several approaches have been proposed for each of the goals we consider, prior work typically focuses on one or two of these applications, while our approach proves effective across all cases.
\item Theoretically, we analyze the effect of our regularization and demonstrate its capability to reduce reliance on spurious features using a Gaussian mixture model setup (Section \ref{sec:th-res}). Additionally, we establish an OOD generalization guarantee in a causal learning framework. Most approaches developed for subgroup robustness or OOD generalization either lack theoretical guarantees or show results tailored to one of these two settings (discussed further in Section \ref{sec:discussion}).

\end{itemize}

\subsection{Related Work}
\label{rel_work}
We discuss related work on simplicity bias, OOD generalization and subgroup robustness, fairness, feature diversification and debiasing methods in this section. A detailed comparison with prior work appears in Section~\ref{sec:discussion}, here we summarize some key differences. A conceptual difference from prior algorithms for these tasks is that we take a more data-driven approach to define spurious features. Hence our approach is more explicit in terms of its assumptions on the data which could make it easier to evaluate and use, and also allows us to prove theoretical guarantees. Various datasets seem to satisfy our assumptions and hence our approach proves effective across varied applications. In addition, in contrast to other related work, it does not involve training multiple complex models, access to group labels, or unlabelled data from the target distribution.
\vspace{-1mm}
\paragraph{Simplicity Bias in NNs.} Several works \citep{geirhos,arpit2017closer,valle2018deep,geirhos2020shortcut,pitfalls,inc-comp,sb1layer,gs} show that NNs trained with gradient-based methods prefer learning solutions which are `simple'. \citet{inc-comp} show that the predictions of NNs trained by SGD can be approximated well by linear models during early stages of training. \citet{sb1layer} show that 1-hidden layer NNs are biased towards using low-dimensional projections of the data for predictions. \citet{geirhos} show that CNNs make predictions which depend more on image texture than image shape. \citet{pitfalls} create synthetic datasets and show that in the presence of simple and complex features, NNs rely heavily on simple features even when both have equal predictive power. In our work, we experiment with datasets in both these papers (in addition to the datasets in Table \ref{tab1}), to investigate the effectiveness of our method to mitigate simplicity bias.  
 \vspace{-1mm}
\paragraph{OOD Generalization.} Towards developing models which perform better in the real world, OOD generalization requires generalization to data from new \emph{environments}. Environments are usually defined based on the correlation between some spurious feature and the label. Various methods aim to recover a predictor that is \emph{invariant} across a set of environments. \citet{irm} develop the invariant risk minimization (IRM) framework where environments are known, while \citet{eiil} propose environment inference for invariant learning (EIIL), to recover the invariant predictor, when the environments are not known. Predict then interpolate (PI) \citep{pi} and BLOOD \citep{blood} use environment-specific ERMs to infer groups based on the correctness of predictions. We compare our approach with all these methods across several datasets to showcase its effectiveness in improving OOD generalization. 
\vspace{-1mm}
\paragraph{Subgroup Robustness.} In many applications, models should not only do well on average but also do well on subgroups within the data. Several methods \citep{gdro,ltf,lastlayer,barack,afr} have been developed to improve the worst-group performance of a model. One widely used approach is to optimize the worst case risk over a set of subgroups in the data \citep{cvardro,gdro,setlur}. CVaRDRO \citep{cvardro} optimizes over all subgroups in the data, which is somewhat pessimistic, whereas GDRO \citep{gdro} does this over a set of predefined groups. However, group knowledge may not always be available, various methods try to identify or infer groups and reweight minority groups in some way, when group labels are not available \citep{ltf,jtt,george} or partially available \citep{barack}. Just train twice (JTT) \citep{jtt} uses ERM to identify the groups based on correctness of predictions. Learning from failure (LfF) \citep{ltf} simultaneously trains two NNs, encouraging one model to make biased predictions, and reweighting the samples it finds harder-to-learn (larger loss) to train the other model. We compare our approach with all these methods across datasets from Table \ref{tab1} to investigate its effectiveness in improving subgroup robustness. 
\vspace{-1mm}
\paragraph{Fairness.} Several works aim to ensure \emph{fair} predictions of models across subgroups in the data, which are defined based on sensitive demographic information. Commonly used notions of group fairness include demographic parity \citep{dem-par}, which aims to ensure that the representation of different demographic groups in the outcomes of a model is similar to their overall representation, and equalized odds \citep{eq-opp}, which aims for equal predictive performance across groups using metrics such as true positive rate. Fairness interventions used when group information is available include: data reweighting to balance groups before training \citep{rewt}, learning separate models for different subgroups \citep{decoupled}, and post-processing of trained models, such as adjusting prediction thresholds based on fairness-based metrics \citep{eq-opp}. 

Various approaches have been proposed to achieve fairness when demographic information may not be available. Multicalibration \citep{multicalib} aims to learn a model whose predictions are calibrated for all subpopulations that can be identified in a computationally efficient way. \citet{replossmin} proposes a distributionally robust optimization (DRO)-based approach to minimize the worst-case risk over distributions close to the empirical distribution to ensure fairness. \citet{lahoti} proposes adversarial reweighted learning (ARL), where an auxiliary model identifies subgroups with inferior performance and the model of interest is retrained by reweighting these subgroups to reduce bias. The authors also develop the Adult-Confounded dataset to evaluate their approach, which we also consider in our work and compare with ARL. \citet{snob} characterizes the notion of social norm bias (SNoB) using an occupation prediction task on the Bios dataset \citep{bios-data}. SNoB is a form of algorithmic unfairness arising from the associations between an algorithm's predictions and individuals' adherence to inferred social norms. We experiment with the Bios data to investigate the effectiveness of our approach in reducing SNoB. 

\vspace{-1mm}
\paragraph{Feature Diversification and De-biasing Methods.} Deep neural networks are known to exhibit unwanted biases. For instance, CNNs trained on image data may exhibit texture bias \citep{geirhos}, and language models trained on certain datasets may exhibit annotation bias \citep{guru-nlp}. Several methods have been proposed to mitigate these biases. \citet{debias} introduce a framework to learn de-biased representations by encouraging them to differ from a reference set of biased representations. \citet{debian} propose to train two models alternatively, using one to identify biases using an equal opportunity violation criterion and training the other with a reweighted cross-entropy loss to make unbiased predictions. \citet{toogood} reduce shortcut reliance by upweighting samples based on the misclassification probability of a simple model to train a complex model. \citet{utama} propose a confidence regularization approach to encourage models to learn from all samples. Recent works also show that deep neural networks tend to amplify the societal biases present in training data \citep{wang-iccv,zhao-nlp} and they propose domain-specific strategies to mitigate such amplification.

Several works aim to improve feature learning for better generalization. \citet{bonsai} aims to learn a shared representation using a succession of training episode, where they train new classifiers in each episode to do better on misclassified samples from previous episodes. \citet{evading-sb} also involves learning a shared representation and training multiple classifiers, where the alignment between their gradients is regularized to encourage feature diversity. \citet{agdisag, divdis} do so by encouraging disagreement between two models leveraging unlabeled samples from the target domain. 
Compared to these approaches, our approach does not involve training multiple complex models, or require access to unlabelled data from the target distribution (discussed in detail in Section \ref{sec:discussion}).
\section{Spurious Features are Simple and Predictive}
\label{sec:define_simple}
In this section, we conduct an experiment to show that surrogate features are generally `simpler' than invariant features (examples in Table \ref{tab1}). First, we define \textit{simple models/features} for a task as follows:

\begin{definition}[Simple models and features] \label{def:simple_models}Consider a  task on which benchmark models which achieve high in-distribution accuracy have a certain complexity (in terms of number of parameters, layers etc.). We consider models that have significantly lower complexity than  benchmark models as \emph{simple models}. Similarly, we consider features that can be effectively learned using simple models as \emph{simple features}.
\end{definition}

Definition \ref{def:simple_models} proposes a metric of simplicity which is task dependent since it depends on the complexity of the models which get high accuracy on the task. We choose to not define quantitative measures in Definition \ref{def:simple_models} since those would be problem dependent. As an example, for colored-MNIST (CMNIST), neural networks with non-linearities are necessary to achieve high accuracy, and for Waterbirds, deeper networks such as ResNet-50 \citep{resnet50} achieve best results. Therefore, a linear model and a shallow CNN can be considered as simple models for these two datasets, respectively. We discuss simple model selection in more detail in Section \ref{sec:cmid}.

\begin{wraptable}[8]{r}{0.75\columnwidth}
\vspace{-2mm}
\centering
\resizebox{0.75\columnwidth}{!}{%
\begin{tabular}{lccccccc}    
\hline 
\multirow{3}{*}{\parbox{0.75cm}{Dataset}}& \multicolumn{4}{c}{ Simple Model } & \multicolumn{2}{c}{ Benchmark Model} \\
& \multicolumn{2}{c}{ Predict surrogate feature } & \multicolumn{2}{c}{ Predict invariant feature} & \multicolumn{2}{c}{ Predict invariant feature}\\
 &   Train & Test   & Train & Test & Train & Test \\
\hline 
CMNIST & $100$ & $100$ & $86.2\pm 0.2$ &
$86.6\pm 0.3$ & $97.3\pm 0.1$ & $96.7\pm0.1$\\
Waterbirds & $79.6\pm 0.6$ & $78.4\pm 0.6$ & $60.5\pm 2.5$ & $60.4\pm 2.4$ & $98.8\pm1.2$ & $96.2\pm1.1$\\
\hline
\end{tabular}}
\caption{Comparison between performance for predicting the simple feature and the complex feature.}
    \label{simple-complex}
\end{wraptable}

Importantly, we observe that these simple models can still achieve good performance when the task is to distinguish between surrogate features even though they are not as accurate in predicting the invariant feature. Specifically, for CMNIST, we compare the performance of a linear model on color classification and digit classification on clean MNIST data. Similarly, for Waterbirds, we consider a shallow CNN (specifically, the \texttt{2DConvNet1} architecture in Appendix Fig.~\ref{ConvNet}) and compare its performance on background classification (using images from the Places dataset \citep{places}) and bird type classification (using segmented images of birds from the CUB dataset \citep{cub}). The results in Table \ref{simple-complex} indicate that surrogate features for these datasets are simple features, since they can be predicted much more accurately by simpler models than invariant features. Unsurprisingly, the respective benchmark models can capture the task-relevant features much more accurately as they are more complex than the simple models considered in each case. 

Motivated by these observations, we define \textit{spurious features} as follows. Operationally, our definition has the advantage that it does not require knowledge of some underlying causal graph or data from multiple environments to determine if a feature is spurious.  

\begin{definition}[Spurious features]\emph{ Spurious features} are simple features that are still reasonably correlated with the target label.
\end{definition}

\begin{wraptable}[8]{r}{0.32\columnwidth}
\vspace{-2mm}
\centering
\resizebox{0.32\columnwidth}{!}{%
\begin{tabular}{lcc}    
\hline 
Dataset &  Train  & Test  \\
\hline 
CMNIST & $84.9\pm 0.2$ &
$10.7\pm 0.3$  \\
Waterbirds & $93.3\pm 0.5$ & $54.9\pm 1.1$ \\
\hline
\end{tabular}}
\caption{Train and test performance of the simple model on the target task.}
    \label{full-task}
\end{wraptable}

In the presence of such features, simplicity bias leads the model to preferring such features over invariant features that are more complex. We also verify that when simple models are trained on the target tasks on these datasets, they tend to rely on these spurious features to make accurate predictions. Specifically, we consider the digit classification task using CMNIST data, where the correlation between the color and the label in the test set is $10\%$, and birdtype classification using Waterbirds data, where the test set consists of balanced groups ($50\%$ correlation). Table \ref{full-task} shows that the test accuracy is close to the correlation between the spurious feature-based group label and the target label. This indicates that simple models trained on the target task utilize the spurious features to make predictions. 

We briefly note here that spurious features as defined by us may not always be irrelevant for the task, and mitigating simplicity bias may not always be desirable (further discussed with other limitations in Section \ref{sec:discussion}). Indeed, through similar experiments, we found that for the CelebA dataset the spurious and task-relevant features are not significantly different in terms of complexity (details in Section~\ref{sec:celeba}). As a result, our approach does not lead to much improvement in the worst-group accuracy on this dataset (Section \ref{sec:subgroup_robustness}). However, it is impossible for any learning rule to generalize across all types of distribution shifts (see Section \ref{sec:discussion} for more discussion); one can only aim to generalize under specific types of distribution shifts by employing the appropriate inductive bias. Reducing reliance on simple, predictive features represents one such assumption, which seems reasonable based on the results presented above and proves effective in several cases, as demonstrated in Section \ref{sec:expts}.
\vspace{-2mm}
\section{CMID: Learning in the Presence of Spurious Features}
\label{sec:cmid}
In this section, we outline our proposed approach to mitigate simplicity bias. Our approach leverages the fact that simple models can capture surrogate features much better than invariant features (Table \ref{simple-complex}), and rely on the spurious features to make predictions, even when trained on the target task (Table \ref{full-task}). Thus, by ensuring that the final model has low conditional mutual information with respect to such a simple model, we encourage it to utilize a more diverse set of features. 

Let $Z=(X,Y)$ denote an input-label pair, where $X\in\mathcal{X},Y\in\{0,1\}$, sampled from some distribution $\mathcal{D}$, $D$ denote a dataset with $n$ samples, $\mdl(\theta):\mathcal{X}\rightarrow [0,1]$ denote a model, parameterized by $\theta$ (shorthand $\mdl$). The predictions of the model are given by $\mathds{1}[\mdl(X)>0.5]$. Subscripts $(\cdot)_s$ and $(\cdot)_c$ denote simple and complex, respectively. Let $\mathcal{\MakeUppercase{\mdl}}$ denote the class of all models $M$, $\ell_{\mdl}(Z):\mathcal{X}\times \{0,1\}\rightarrow \R$ denote a loss function, and $\sigma(x,T) = \frac{1}{1 + e^{-Tx}}$ denote the sigmoid function, with temperature parameter $T$.

Let $H(\cdot)$ denote the Shannon entropy of a random variable. $I(\cdot;\cdot)$ measures Shannon mutual information between two random variables. With slight abuse of notation, let $\mtil$ and $\ytil$ also denote the (binary) random variables associated with the predictions of model $\mdl$ on datapoints $X_i$'s and the labels $Y_i$'s, across all $i \in[n]$, respectively. The conditional mutual information (CMI) between the outputs of two models given the label is denoted by $I(\mtil_1;\mtil_2|\ytil)$. 

\paragraph{}
The empirical risk minimizer (ERM) for class $\mathcal{\mdl}$ is denoted as $M^*$, and is given by:

\begin{align*}
    \mdl^*:={\text{ERM}}(\mathcal{\MakeUppercase{\mdl}})=\argmin_{\mdl\in \mathcal{\MakeUppercase{\mdl}}}\frac{1}{n}\sum\limits_{i\in[n]}\ell_{\mdl}(Z_i)=\argmin_{\mdl\in \mathcal{\MakeUppercase{\mdl}}}\E_D\,\ell_{\mdl}(Z).
\end{align*}

We consider the class of simple models $\mathcal{\mdl}_s\subset \mathcal{\mdl}$ and complex models $\mathcal{\mdl}_c=\mathcal{\mdl}\setminus \mathcal{\mdl}_s$. We now describe our proposed two-stage approach CMI Debiasing (CMID):
\begin{itemize}[leftmargin=*]
    \item First, learn a simple model $\mdl_s$, which minimizes risk on the training data:
    \begin{align*}
        \mdl_{s}^*=\text{ERM}(\mathcal{\mdl}_s).
    \end{align*}
    
    \item Next, learn a complex model $\mdl_{c}$ by regularizing its CMI with $\mdl_{s}$.
    \begin{align*}
        \mdl_{c}=\argmin_{\mdl\in \mathcal{\MakeUppercase{\mdl}}}\E_D\,\ell_{\mdl}(Z)+\lambda \hat{I}_D(\mtil;\mtil_s^*|\ytil),
    \end{align*}
    where $\hat{I}_D(\mtil;\mtil_s^*|\ytil)$ denotes the estimated CMI over $D$, and $\lambda$ is the regularization parameter.
\end{itemize}

A few remarks are in order about the choice of our regularizer and the methodology to select the simple model class. Note that we penalize the conditional mutual information instead of mutual information. This is because both $M_s^*$ and $M$ are expected to have information about $Y$ (for e.g. in the Waterbirds dataset both bird type and background type are correlated with the label). Hence they will not be independent of each other, but they are closer to being independent when conditioned on the label. We also note that we use conditional mutual information to measure dependence, instead of other measures such as enforcing orthogonality of the predictions. This is for the simple reason that  mutual information measures all---potentially non-linear---dependencies between the random variables.

We note that while several models can be considered simple based on Definition \ref{def:simple_models}, the choice of the simple model class $\mathcal{M}_s$ used for our approach is task dependent. Intuitively, we want to use the simplest model that we expect to do reasonably well on the given task. Models that are too simple may not be able to capture surrogate features effectively, whereas models that are very complex may rely on task-relevant features, even though such reliance may be weak due to simplicity bias. 

\begin{wrapfigure}[14]{r}{0.55\columnwidth}
  \centering
  \vspace{-3mm}
    \includegraphics[height=110pt]{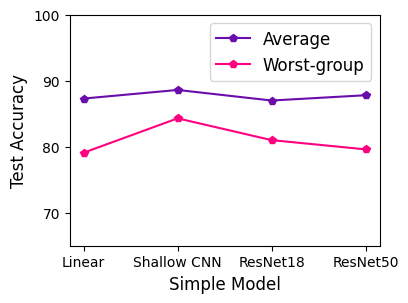} \includegraphics[height=110pt]{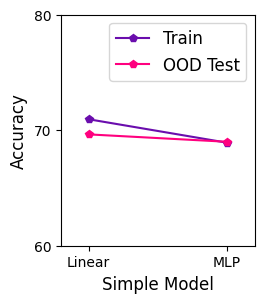}   
    \caption{Performance of CMID using different simple model architectures for Waterbirds (left) and CMNIST (right) dataset. 
    }
    \label{fig:smw-cmid}
  
\end{wrapfigure}
As an example, Fig. \ref{fig:smw-cmid} (left) considers the problem of subgroup robustness on the Waterbirds dataset (where typically the hardest groups are images of waterbirds on land backgrounds and vice versa), and shows a comparison of the average and worst-group accuracy of the model learned by CMID using various simple model architectures (see Section \ref{sec:subgroup_robustness} for more details about the task). We consider four models: a linear model, the shallow CNN (\texttt{2DConvNet1}) used in Table \ref{simple-complex}, a ResNet-18 pretrained on ImageNet and the ResNet-50 pretrained on ImageNet, which is also the architecture of the benchmark model. We observe that using a shallow CNN as the simple model is most effective for CMID. This is because a linear model is too simple to capture the surrogate feature, i.e. the background in this case, whereas deep models like ResNet are complex enough to learn both types of features, while relying more strongly on the surrogate feature. As a result, regularizing the CMI with respect to such models may not be as effective in reducing reliance on spurious features. Similarly, Fig. \ref{fig:smw-cmid} (right) shows a comparison of train and OOD test when using a linear model and an MLP as the simple model for the CMNIST dataset.

Although the choice of $\mathcal{M}_s$ may impact the performance of CMID, we use a fairly simple selection rule throughout our experiments, which works well. In general, for datasets where the final model is a shallow NN, we used a linear model as the simple model. For image datasets where the final model is a ResNet or DenseNet-based model, we consider shallow 2D CNNs as simple models. For language datasets where BERT-based models are the final models, we consider shallow MLPs or 1D CNNs as simple models. Details about the model architectures can be found in Appendix  \ref{simple-models}. 
\paragraph{}

Next, we discuss how we estimate the CMI, which is used as a regularizer for our approach. To estimate CMI, first consider the conditional (joint) distributions over $n$ samples:

\begin{align*}    p(\mtil\!=\!\MakeLowercase{\mdl}|\ytil\!=\!y)&=\frac{\Sigma_{i\in[n]}\ind[Y_i\!=\!y]\,\zeta(\mdl(X_i),\MakeLowercase{\mdl})}{\Sigma_{i\in[n]}\ind[Y_i\!=\!y]}, \\    p(\mtil\!=\!\MakeLowercase{\mdl},\mtil_s\!=\!\MakeLowercase{\mdl}'|\ytil\!=\!y)&=\frac{\Sigma_{i\in[n]}\ind[Y_i\!=\!y]\,\zeta(\mdl(X_i),\MakeLowercase{\mdl})\,\zeta(\mdl_s(X_i),\MakeLowercase{\mdl}')}{\Sigma_{i\in[n]}\ind[Y_i\!=\!y]},
\end{align*}

where $\MakeLowercase{\mdl},\MakeLowercase{\mdl}',y\in\{0,1\}$, $\zeta(\mdl(X_i),1)\!=\!\ind[\mdl(X_i)\!>\!0.5]$, and $\zeta(\mdl(X_i),0)\!=\!1\!-\!\ind[\mdl(X_i)\!>\!0.5]$. 

Note that the CMI computed using these is not a differentiable function of the model outputs as these densities are computed by thresholding the model outputs. Since we want to add a CMI penalty as a regularizer to the ERM objective and optimize the proposed objective using standard gradient-based methods, we need a differentiable version of CMI. Thus, for practical purposes, we use an approximation of the indicator function $\ind[x>0.5]$, given by $\sigma(x-0.5, T)$, where $T$ determines the degree of smoothness or sharpness in the approximation. 

This can be easily generalized for multi-class classification with $C$ classes. In that case, $\MakeLowercase{\mdl},\MakeLowercase{\mdl}',y\in\{0,\cdots,C-1\}$, $M(X_i)$ is a $C$-dimensional vector with the $m^{\text{th}}$ entry indicating the probability of predicting class $m$, and $\zeta(\mdl(X_i),m)\!=\ind[\argmax_{j\in[C]} M_j(X_i)=m]$. To make this differentiable, we use the softmax function with temperature parameter $T$ to approximate the indicator function. 

Using these densities, the estimated CMI is:
\begin{align}   
\label{cmi-est-eq}\hat{I}_D(\mtil,\mtil_s|\ytil)\!=\!\sum\limits_{y}p(\ytil\!=\!y_i)\sum\limits_{\MakeLowercase{\mdl},\MakeLowercase{\mdl}'}
p(\mtil=\MakeLowercase{\mdl},\mtil_s\!=\!\MakeLowercase{\mdl}'|\ytil\!=\!y)\log\left[\frac{
p(\mtil\!=\!\MakeLowercase{\mdl},\mtil_s\!=\!\MakeLowercase{\mdl}'|\ytil\!=\!y)}{
p(\mtil\!=\!\MakeLowercase{\mdl}|\ytil\!=\!y)
p(\mtil_s\!=\!\MakeLowercase{\mdl}'|\ytil\!=\!y)}\right].
\end{align}

This estimate is differentiable, making it compatible with gradient-based methods. Therefore, we utilize it as a regularizer for the proposed approach. We include comparison between the estimated CMI and the original CMI (computed with discretized model outputs) in Section~\ref{A1}.
\vspace{-2mm}
\section{Theoretical Results}
\label{sec:th-res}
In this section, we analyze the effect of CMI regularization and show it leads to reduced dependence on spurious features in a Gaussian mixture model. We also obtain a simple OOD generalization guarantee for our approach in a causal learning framework.
\subsection{Effect of CMI Regularization for Gaussian Features}
\begin{figure}[h!]
    \centering
    \includegraphics[height=105pt]{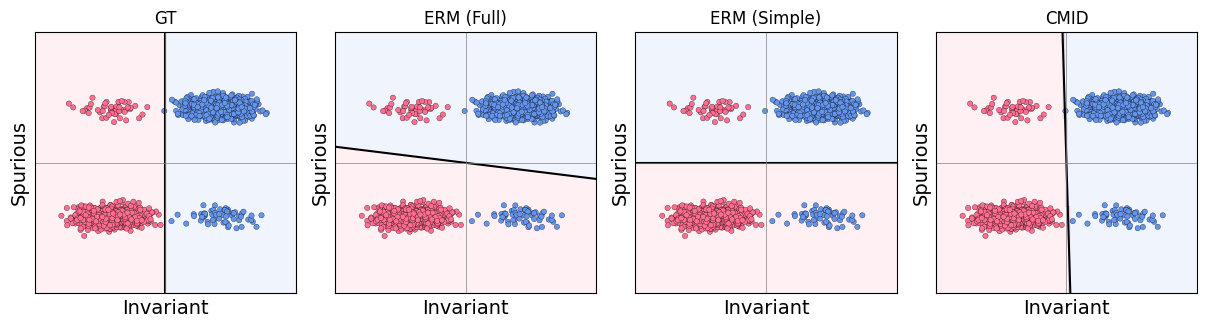}  
    \caption{Results on synthetic Gaussian data generated as per Assumption \ref{assumpt:gauss}, with $2000$ samples, $\ma\!=\!\mb\!=\!5$, $\sia\!=\!1.5,\sib\!=\!0.5$. Left to right: Ground truth (GT) predictor, ERM with $\mathcal{\mdl}$ as the class of linear models, ERM with $\mathcal{\mdl}$ as the class of threshold functions, ERM with CMI constraint, with $\eps\!=\!0.01$.}
    \label{fig:gauss}
\end{figure}

We consider data generated from the following Gaussian mixture model (refer to the left-most panel 
 in Fig. \ref{fig:gauss} for an example of data generated from this distribution).
\begin{assumption}\label{assumpt:gauss}
Let the label $y\sim \mathcal{R}(0.5)$, where $\calR(p)$ is a $\{\pm1\}$ random variable which is 1 with probability $p$. Consider an \emph{invariant} feature $\fta$ and a \emph{spurious} feature $\ftb$, with  distributions:
\begin{equation*}
    \fta \sim \mathcal{N}(y\ma,\sia^2) \text{ and } \ftb \sim \mathcal{N}(a\mb,\sib^2),\label{eq:gauss_setup}
\end{equation*}
where $a\sim y\mathcal{R}(\eta)$ is a spurious attribute, with an unstable correlation with $y$, and $\ma,\mb>0,\eta>0.5$. Assume $\fta\perp\ftb|y$ and let $\mb'=(2\eta-1)\mb$.
\end{assumption}

We note that \citet{gauss1,rosenfeld} consider a similar Gaussian mixture based data model for core and spurious features. Intuitively, the mean of a feature determines the margin of the predictor that relies on that feature, while the variance of the feature determines the best possible accuracy for that predictor. 

We consider linear models to predict $y$ from the features $X_1$ and $X_2$. Let $\calM=\{(w_1,w_2): w_1,w_2 \in \R\}$ be all possible linear models and $\calM_s=\{(0,w): w \in \R\}\cup \{(w,0): w \in \R\}$ be a simpler model class which only uses one of the two features. We consider the mean squared error (MSE) loss, and the ERM solution is given by:
\begin{align*}
    \text{ERM}(\calM)=\argmin_{\w\in\calM }\,\E\,(\wa\fta+\wb\ftb-y)^2.
\end{align*}

\begin{propo}\label{prop:erm_gaussian}
    ERM($\calM$) satisfies $\tfrac{\wa}{\wb}\!=\!\tfrac{\ma}{\mb'}\tfrac{\sib'^2}{\sia^2}$. When  $\tfrac{\ma}{\mb'}\tfrac{\sib'^2}{\sia^2}\!<\!1$, ERM($\calM_s$)$=\!\left[0,\tfrac{\mb'}{\sib^2}\right]$ (upto scaling). 
\end{propo}

We now consider the effect of CMI regularization. Since both the features are of similar complexity here, we consider the core feature to have a lower signal-to-noise ratio, i.e. $\tfrac{\ma}{\mb'}\tfrac{\sib'^2}{\sia^2}\!<\!1$, to model the observation that the spurious feature is learned more easily with ERM. Then, as per Proposition~\ref{prop:erm_gaussian} in the first step we learn a simple model $w_2^*X_2$ which uses only $X_2$. We now consider the ERM problem but with a constraint on the CMI:
\begin{align}
\label{cmi-gauss}
    \text{ERM}_{\calC}(\calM)=\argmin_{\w\in\calM }\,\E\,(\wa\fta+\wb\ftb-y)^2
    \text{ s.t. } I(\wa\fta+\wb\ftb;\wb^*\ftb|y)\leq \nu.  \end{align}

\vspace{-3mm}

We show the following guarantee on the learned model.
\begin{theorem}
\label{thm:gauss1}
Let data be generated as per Assumption \ref{assumpt:gauss}. 
For $\nu=0.5\log(1+\eps^2)$ for some $\eps$:

1. When $\tfrac{\ma}{\mb'}\tfrac{\sib'^2}{\sia\sib}>\tfrac{1}{\eps}$, the solution to \eqref{cmi-gauss} 
is the same as ERM($\calM$), so $\tfrac{\wa}{\wb}=\tfrac{\ma}{\mb'}\tfrac{\sib'^2}{\sia^2}$.

2. Otherwise, $\wa$ is upweighted and the solution to \eqref{cmi-gauss} 
satisfies $\tfrac{|\wa|}{|\wb|}=\tfrac{1}{\eps}\tfrac{\sib}{\sia}$. 
\end{theorem}

\begin{wrapfigure}[18]{r}{0.62\columnwidth}
    \centering
    \vspace{-3mm}
    \includegraphics[height=120pt]{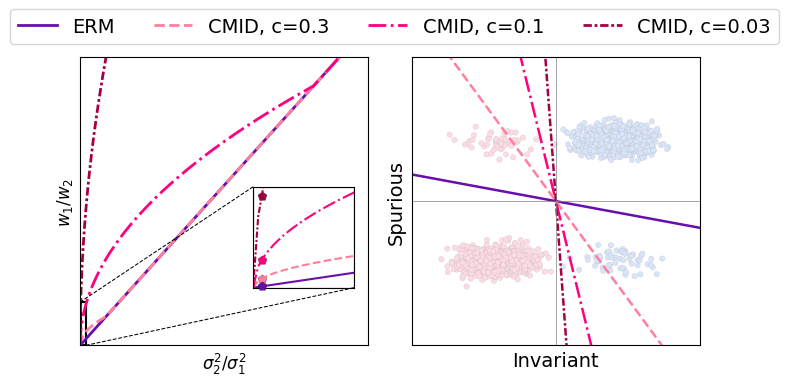}  \caption{Effect of CMI regularization (wrt $\ftb$)  for different values of $c$ (corresponding to regularization strength). Left: Lower values of $\eps$ indicate stronger CMI regularization, resulting in more upweighting of $\wa$ wrt $\wb$. Inset shows a zoomed-in region and markers compare the three solutions when $\sib^2/\sia^2=1/6$. Right: Decision boundaries for predictors corresponding to the markers. 
    }
    \label{fig:reg-plot}
\end{wrapfigure}
Theorem \ref{thm:gauss1} suggests that for an appropriately small $\eps$, regularizing CMI with the simple model leads to a predictor which mainly uses the invariant feature. This is supported by experimental results on data drawn according to Assumption \ref{assumpt:gauss}, shown in Fig.~\ref{fig:gauss}.

In Fig. \ref{fig:reg-plot}, we visualize the relationship between $\tfrac{\wa}{\wb}$ and $\tfrac{\sib^2}{\sia^2}$ predicted in Theorem \ref{thm:gauss1} (assuming $\ma\!=\!\mb$ and $\eta\!=\!0.95$.) We see that a lower value of $\eps$ promotes conditional independence with $X_2$ and upweighs $\wa$ more strongly. When $\eps\rightarrow 0$,  $\wb\rightarrow 0$. 

Next, we consider the case where there are multiple spurious features, by adding another feature 
\begin{align}
    \ftc\sim\mathcal{N}(a'\mc,\sic^2), \text{ where  }a'\sim y \calR(\eta')\label{eq:x3} 
\end{align}
to the setup in Assumption \ref{assumpt:gauss}. We show that regularizing CMI with respect to the optimal predictor that uses $\ftb$ and $\ftc$ also results in $\wa$ being upweighted. We consider the constrained problem:
\begin{align}
\label{cmi-gauss2}
    \text{ERM}_{\calC}(\calM)=\argmin_{\w\in\calM }\,& \,\E\,(\wa\fta+\wb\ftb+\wc\ftc-y)^2\nonumber\\
    \text{ s.t. } &I(\wa\fta+\wb\ftb+\wc\ftc;\wb^*\ftb+\wc^*\ftc|y)\leq \nu,   
\end{align}
where $(\wb^*,\wc^*)\propto\left({\mb'}/{\sib'^2},{\mc'}/{\sic'^2}\right)$ (the weights for the optimal linear model which uses $X_2$ and $X_3$). We show  the following result.

\begin{theorem}
\label{gauss-th-res2}

Let the data be generated as per Assumption \ref{assumpt:gauss} and \eqref{eq:x3}. Let $\nu\!=\!0.5\log(1+\eps^2)$ for some sufficiently small $\eps$, and $\tfrac{\sib^2}{\sib'^2}=\tfrac{\sic^2}{\sic'^2}$. Then the solution to \eqref{cmi-gauss2} satisfies: $
\tfrac{|\wb\mb'+\wc\mc'|}{|\wa|\sia}= \eps'$, where $\eps'=2\eps\sqrt{\tfrac{\mb'^2}{\sib^2}+\tfrac{\mc'^2}{\sic^2}}$. Moreover, 
$\wb,\wc\propto\eps'\wa$.
\end{theorem}

Similar to before, the result says that for a sufficiently small CMI constraint we learn a model which mainly uses the invariant feature. 
\vspace{-2mm}
\subsection{OOD Generalization in a Causal Learning Framework}

Following the setting in \citet{irm,hrm}, we consider a dataset $D = \{D^e\}_{e\in\calE_{tr}}$, which is composed of data $D^e \sim \calD_e^{n_e}$ gathered from different training environments $e \in \calE_{tr}$, where $e$ denotes an environment label, $n_e$ represents the number of samples in $e$. $\calE_{tr}$ denotes the set of training environments.

The problem of finding a predictor with good OOD generalization performance, can be formalized as:
\begin{align*}
\argmin_{\mdl\in\mathcal{\mdl}} \max_{e\in \calE }
\E_D[\ell_\mdl(Z)|e],
\end{align*}
\textit{i.e.}, optimizing over the worst-case risk on all environments in set $\calE$. Usually, $\calE \supset \calE_{tr}$, and hence, the data and label distribution can differ significantly for $e\in\calE_{tr}$ and $e\in\calE\setminus \calE_{tr}$. This makes the OOD generalization problem hard to solve.

The invariant learning literature assumes the existence of invariant and variant features. In this section, we assume that the model of interest, say $M(X)$ is composed of a featurizer $\Upphi$ and a classifier $\omega$ on top of it, \textit{i.e.} $M(X)=\omega\circ\Upphi(X)$. For simplicity, we omit the argument $X$ and assume that learning a featurization includes learning the corresponding classifier, so we can write $M=\Upphi$. Let $E$ be a random variable sampled from a distribution on $\calE$.

\begin{definition}[Invariant and Variant Predictors]
    A feature map $\Upphi$ is called \emph{invariant} and is denoted by $\Phi$ if $Y\perp E| \Phi$, whereas it is called \emph{variant} and is denoted by $\Psi$ if $Y\not\perp E| \Psi$.
\end{definition} 
 
Several works \citep{irm,hrm,eiil} attempt to recover the invariant feature map by proposing different ways to find the maximally invariant predictor \citep{chang-ir}, defined as: 
\begin{definition}[Invariance Set and Maximal Invariant Predictor]\label{mip}
The invariance set $\calI$ with respect to environment set $\calE$ and hypothesis class $\mathcal{\mdl}$ is defined as:
$$\calI_\calE(\mathcal{\mdl}) =\{\Phi:Y \perp E|\Phi\}
= \{\Phi : H[Y |\Phi] = H[Y |\Phi, E]\}.$$
The corresponding maximal invariant predictor (MIP) of $\calI_\calE(\mathcal{\mdl})$ is
$\Phi^*= \argmax_{\Phi\in \calI_\calE(\mathcal{\mdl})} I(Y ; \Phi) $.
\end{definition}
The MIP is an invariant predictor that captures the most information about $Y$. Invariant predictors guarantee OOD generalization, making MIP the optimal invariant predictor (Theorem 2.1 in \citet{hrm}).

As discussed in Section \ref{sec:intro}, most current work assumes that the environment labels $e$ for the datapoints are known. However, environment labels typically are not provided in real-world scenarios. In this work, we do not assume access to environment labels and instead, we rely on another aspect of these features: are they simple or complex? We formalize this below:

\begin{wrapfigure}[11]{r}{0.3\textwidth}
    \centering
    \vspace{-1mm}
    \includegraphics[height=75pt]{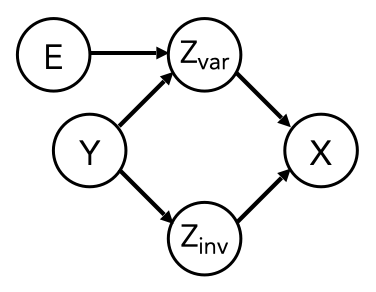}  
    \caption{Our causal model.  Latent  variables $\text{Z}_{\text{inv}}$ and $\text{Z}_{\text{var}}$ correspond to {invariant} and {variant} features  $\Phi^*$ and $\Psi^*$  respectively.}
    \label{fig:ucm}
\end{wrapfigure}

\begin{assumption}[Simple and Complex Predictors]
    \label{simple-complex-fts}The invariant feature comprises of complex features, \textit{i.e.} $\Phi^*=[\Phi_c]$, where 
    $\Phi_c\in\mathcal{\mdl}_c$. 
     The variant feature comprises of simple and complex features, \textit{i.e.}, $\Psi^*=[\Psi_c,\Psi_s]$, where $\Psi_c\in\mathcal{\mdl}_c$, $\Psi_s\in\mathcal{\mdl}_s$ and $I(Y;\Psi_s)>0$.
\end{assumption}

We consider the underlying causal model in \citet{rosenfeld} which makes the following assumption.

\begin{assumption}[Underlying Causal Model] \label{ucm-asm} Given the model in Fig.~\ref{fig:ucm}, $I(\Phi,\Psi|Y)\!=\!0$ and $I(\Psi_s,\Psi_c|Y)\!>\!I(\Psi_s,\Psi_c|Y,E)$.
\end{assumption}

The following simple result shows that our method finds the maximal invariant predictor, and thus generalizes OOD.

\begin{propo}\label{th:ood}
Let $\text{ERM}(\mathcal{\mdl}_s)=M_s^*$. 
    Under Assumptions \ref{simple-complex-fts} and \ref{ucm-asm}, the solution to the problem:
    \begin{align}
    \label{ood-th-eqn}
        \argmin_{\mdl\in\mathcal{\mdl}}\E \, \ell_\mdl(Z) \text{ s.t. } I(\mdl;M_s^*|Y)=0
    \end{align}
    is $\mdl=\Phi^*$, the maximal invariant predictor.
\end{propo}

Here, we note that Assumption \ref{simple-complex-fts} is key to our result. Although this assumption may not always hold directly in practice, it establishes one condition under which our approach can recover the invariant predictor, even though it does not have an explicit causal/invariant learning motivation or access to environment labels.
\vspace{-2mm}
\section{Experiments}
\label{sec:expts}
We show that \name reduces simplicity bias, and yields improvements across various robustness, OOD generalization and fairness metrics. We first show that \name mitigates simplicity bias in Slab and ImageNet-9 datasets. We then evaluate \name on various OOD generalization tasks. These include data with multiple spurious features, real-world medical data, and a fairness task. Finally, we test CMID on some benchmark datasets for subgroup robustness and a fairness application\footnote{Our code is available at \url{https://github.com/estija/CMID}.}. We note that past approaches usually target one or two of these problem settings. Thus, in each section we generally choose the most task-relevant methods to compare with, as established in prior work on that task. 
For consistency, in addition to ERM, we compare with JTT on most of the tasks, since it is the most similar method to CMID in terms of its requirements. Throughout, we observe that CMID improves considerably on ERM and compares favorably with JTT and domain-specific approaches on most data.
\subsection{Mitigating Simplicity Bias}
In this section, we show that \name mitigates simplicity bias and encourages the model to learn more diverse features on two tasks. These include a synthetic Slab dataset, that was used to demonstrate extreme simplciity bias in NNs \citep{pitfalls}, and ImageNet-9 dataset, where CNNs are known to rely more on texture rather than shapes to make predictions \citep{geirhos}.
\paragraph{Slab Data.}
\begin{wrapfigure}[14]{r}{0.52\textwidth}
    \centering
    \vspace{-4mm}
\includegraphics[height=120pt]{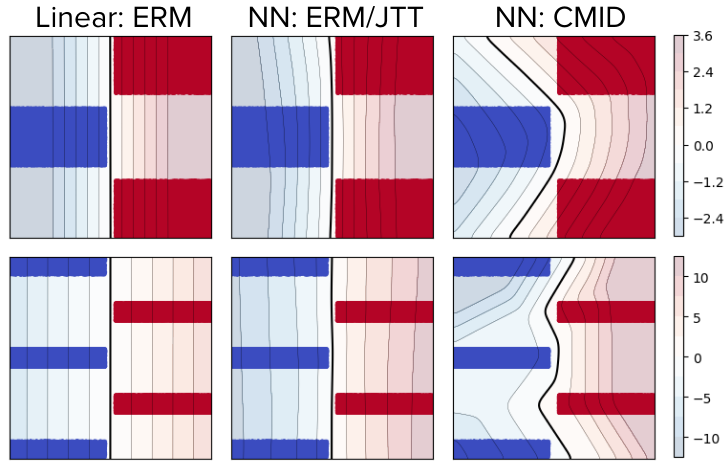}  
    \caption{Results on the slab dataset, when training a linear model and a 1-hidden-layer NN using ERM, JTT, and CMID.}
    \label{fig:slab}
\end{wrapfigure}
Slab data was proposed in \citet{pitfalls} to model simplicity bias. Each feature is composed of $k$ data blocks or slabs. We consider two configurations of the slab data, namely 3-Slab and 5-Slab, as shown in Fig.~\ref{fig:slab}. In both the cases, the first feature is linearly separable. The second feature has 3 slabs in the 3-Slab data and 5 slabs in the 5-Slab data. The first feature is simple since it is linearly separable, while the features with more slabs involve a piece-wise linear model and are thus complex. The linear model is perfectly predictive, but the predictor using both types of features attains a much better margin, and generalizes better under fixed $\ell_1$-norm perturbations to the features. Fig. \ref{fig:slab} shows the decision boundary using ERM/JTT and the proposed approach. We see that CMID encourages the model to use both features and attain better margin. We note that since both the features are fully predictive in this setting, approaches like JTT are ineffective as they rely on incorrect predictions of one model to train another model by upweighting such samples.

\paragraph{Texture vs Shape Bias on ImageNet-9.}
\citet{geirhos} showed that CNNs trained on ImageNet tend to make predictions based on image texture rather than image shape. To quantify this phenomenon, the authors designed the GST dataset, which contains synthetic images with conflicting shape and texture (e.g., image of a cat modified with elephant skin texture as a conflicting cue). The \emph{shape bias} of a model on the GST dataset is defined as the number of samples for which the model correctly identifies  shape compared to the total number of samples for which the model correctly identifies either shape or texture. 

\begin{wraptable}[10]{r}{0.4\columnwidth}
\vspace{-2mm}
    \centering
    \resizebox{0.4\columnwidth}{!}{%
\begin{tabular}{cccc}
\hline ImageNet-9 Class & GST Dataset Class\\
\hline dog & dog \\
bird & bird\\ 
wheeled vehicle & bicycle,  vcar, truck\\
carnivore & bear, cat\\
musical instrument & keyboard\\
\hline
\end{tabular}
}\caption{ImageNet-9 classes mapped to corresponding GST dataset classes.}
\label{mapping}
\end{wraptable}

We consider ImageNet-9, a subset of ImageNet organized into nine classes by \citet{imagenet9}. Each of the nine condensed classes consists of images from multiple ImageNet classes. These include dog, bird, wheeled vehicle, reptile, carnivore, insect, musical instrument, primate, and fish. To interpret a prediction from a model trained on ImageNet-9 as a GST dataset prediction, we consider a subset of classes from both and use the mapping listed in Table \ref{mapping}. Specifically, to determine whether a model trained on ImageNet-9 predicts correctly on a GST image, we first determine which of the 5 ImageNet-9 classes from Table \ref{mapping} has the highest probability based on the model's output, and then use the mapping to obtain the predicted GST class label.

Thus, the shape bias is calculated as:\footnote{We use the code by \citet{geirhos} available at \url{https://github.com/rgeirhos/texture-vs-shape}.}
\[\text{shape bias} = \frac{\text{number of correct shape predictions}}{\text{number of correct shape predictions} + \text{number of correct texture predictions}}\times 100.\]

\begin{wraptable}[5]{r}{0.35\columnwidth}
    \centering
    \vspace{-3mm}
\resizebox{0.32\columnwidth}{!}{%
        \begin{tabular}{ccc>{\columncolor{magenta!10}}c}
        \hline 
        Method & ERM & JTT 
        &  \name \\
        \hline
        Shape Bias & $38.6$ & $45.8$ & $\mathbf{51.8}$ \\
        \hline
        \end{tabular}
    }
    \caption{Comparison of shape bias on ImageNet-9 dataset.}
    \label{imnet}
\end{wraptable}
A shape bias of 100 indicates that the model always uses shape when shape conflicts with texture, whereas 0 indicates that the model always uses texture instead of shape. 

We train a ResNet50 model using ERM and CMID. In Table~\ref{imnet}, we show that compared to ERM, training with \name mitigates texture bias and encourages the model to rely on shape for predictions. 
\vspace{-2mm}
\subsection{Better OOD Generalization}
In this section, we show that \name leads to improved OOD generalization on several datasets. We first evaluate our approach on the widely used CMNIST data \citep{irm}, where color is a spurious feature, as well as a recent variant with an additional spurious feature in the form of patch, CPMNIST \citep{blood}. We also consider the Camelyon17, a real-world medical image dataset from the WILDS benchmark \citep{wilds}, and the Adult-Confounded dataset \citep{eiil}, a semi-synthetic variant of the UCI-Adult dataset,  which is a well-known fairness benchmark.
\paragraph{Synthetic: CMNIST and Color+Patch MNIST.}
We present results on two variants of the MNIST dataset \citep{mnist-cite}, which contains images of handwritten digits, using a binary digit classification task ($<5$ or not). The colored-MNIST data was proposed in \citet{irm}, where color (red/green) is injected as a spurious feature, with unstable correlation $1-p_e$ with the label across environments. 
\begin{table}[h!]
\centering
\resizebox{0.95\columnwidth}{!}{\begin{tabular}{lccccccc}    
\hline &\multirow{3}{*}{\parbox{0.9cm}{Group labels}}& \multicolumn{3}{c}{ Bias: Color } & \multicolumn{3}{c}{ Bias: Color \& Patch } \\
Method &  & Test (i.i.d) & Test (OOD) & $\delta_{g a p}$ & Test (i.i.d) & Test (OOD) & $\delta_{g a p}$ \\
& & $p_e=0.1$ & $p_e=0.9$ & & $p_e=0.1$ & $p_e=0.9$ & \\
\hline ERM &No& $88.6_{\pm 0.3}$ & $16.4_{\pm 0.8}$ & $-72.2$ & $93.7_{\pm 0.3}$ & $14.0_{\pm 0.5}$ & $-79.7$ \\
EIIL \citep{eiil} 
&No& $71.7_{\pm 1.6}$ & $62.8_{\pm 5.0}$ & $-8.9$ & $65.3_{\pm 8.4}$ & $53.0_{\pm 5.6}$ &$-12.3$ \\
JTT \citep{jtt} 
&No& $72.2_{\pm 1.1}$ & $\underline{64.6}_{\pm 0.56}$ & $-7.6$ & $64.0_{\pm 2.7}$ & $\underline{56.2}_{\pm 2.7}$ &$-7.8$ \\
\rowcolor{magenta!10}
\name &No& $69.2_{\pm 0.9}$ & $\mathbf{68.9}_{\pm 0.9}$ & $-0.3$ & $60.3_{\pm 2.7}$ & $\mathbf{59.4}_{\pm 1.0}$ &$-0.9$ \\
\hline
IRM \citep{irm} 
&Yes& $71.4_{\pm 0.9}$ & $66.9_{\pm 2.5}$ & $-4.5$ & $93.5_{\pm 0.2}$ & $13.4_{\pm 0.3}$ & $-80.1$ \\
GDRO \citep{gdro} 
&Yes& $89.2_{\pm 0.9}$ & $13.6_{\pm 3.8}$ & $-75.6$ & $92.3_{\pm 0.3}$ & $14.1_{\pm 0.8}$ & $-78.2$ \\
PI \citep{pi} 
&Yes& $70.3_{\pm 0.3}$ & $70.2_{\pm 0.9}$ & $- 0 . 1$ & $85.4_{\pm 0.9}$ & $15.3_{\pm 2.7}$ & $-70.1$ \\
BLOOD \citep{blood} 
&Yes& $70.5_{\pm 1.1}$ & $7 0 . 7_{\pm 1 . 4}$ & $0 . 2$ & $68.3_{\pm 2.3}$ & $6 2 . 3_{\pm 3 . 3}$ & $- 6 . 0$ \\
\hline Optimal &-& $75$ & $75$ & $0$ & $75$ & $75$ & $0$ \\
\hline
\end{tabular}}
\caption{Comparison of average test accuracies on i.i.d and OOD data and their difference ($\delta_{gap}$) on Colored MNIST and Color+Patch MNIST datasets. \textbf{Bold} and \underline{underlined} numbers indicate the best and second-best OOD performance among the methods that don't use group labels.} 
    \label{cmnist}
\end{table}

The train data has two environments with $p_e=0.1,0.2$ while the test data has $p_e=0.9$, to test OOD performance. Further, it contains $25\%$ label noise to reduce the predictive power of the task-relevant feature: digit shape. We also consider the color+patch MNIST data proposed in \citet{blood}, where an additional spurious feature is injected into the data, in the form of a $3\times 3$ patch. The position of the patch (top left/bottom right) is correlated with the label, with the same $p_e$, but independent of the color.

Following \citet{blood}, we evaluate on i.i.d test data with $p_e=0.1$ and OOD data with $p_e=0.9$ for both the cases (details in the Appendix). Table \ref{cmnist} shows that \name gets competitive OOD performance with methods that require group knowledge, and has the lowest gap $\delta_{gap}$ between test performance on i.i.d and OOD samples, even in the presence of multiple spurious features.
\paragraph{Medical: Camelyon17-WILDS.}
Camelyon17-WILDS is a real-world medical image dataset of data collected from five hospitals \citep{camelyon, wilds}. Three hospitals comprise the training set, one is the validation set and the third is the OOD test set. Images from different hospitals vary visually. The task is to predict whether or not the image contains tumor tissue, and the dataset is a well-known OOD generalization benchmark \citep{blood, wilds}. Table \ref{camelyon} shows that \name leads to higher average accuracies than existing group-based methods when evaluated on images from the test hospital. While JTT attains a comparable test accuracy, it exhibits a considerable drop in the train accuracy.
\begin{table}[h!]
\centering
\resizebox{0.8\columnwidth}{!}{%
\begin{tabular}{ccccccc>{\columncolor{magenta!10}}c}
\hline Method &ERM & IRM 
& GDRO 
&  PI 
& BLOOD 
& JTT 
& \name \\
\hline Train Acc & $97.3_{\pm 0.1 }$ & $97.1_{\pm0.1} $ & $96.5_{\pm1.4}  $ &  $93.2_{\pm0.2}  $ & $93.0_{\pm1.8}$ & $88.4_{\pm 1.3}$ & $94.0_{\pm 2.0}$\\
 OOD Test Acc  & $66.5_{\pm4.2  }$ & $59.4_{\pm3.7 } $ & $70.2_{\pm7.3}$ & $71.7_{\pm7.5}  $  & $74.9_{\pm5.0}$ & $\mathbf{78.0}_{\pm6.3}$ & $\underline{77.9}_{\pm 7.7}$\\
\hline
\end{tabular}%
}
\caption{Comparison of average train and OOD test accuracies on Camelyon17-WILDS dataset. }
    \label{camelyon}
\end{table}

\vspace{-4mm}

\paragraph{Fairness: Adult-Confounded.}
The Adult-Confounded dataset is a semi-synthetic variant of the UCI Adult dataset \citep{uci1,uci2}, developed by \citet{eiil}. The UCI Adult dataset contains attributes based on demographics and employment information and the target label a binarized income indicator (thresholded at $\$50, 000$). The task is commonly used as an algorithmic fairness benchmark.

\citet{lahoti,eiil} define four sensitive subgroups based on binarized sex (Male/Female) and race (Black/non-Black) labels: Non-Black Males (G1), Non-Black Females (G2), Black Males (G3), and Black Females (G4). They observe that each subgroup has a different correlation strength with the target label ($p(y=1|G)$), and thus, in some cases, subgroup membership alone can be used to achieve low error rate in prediction. Based on this observation, \citet{eiil} create a semi-synthetic variant of the UCI Adult data, known as Adult-Confounded, where they exaggerate the spurious correlations in the original data. As G1 and G3 have higher values of $p(y=1|\text{G})$ across both the splits, compared to the other subgroups (see \citet{eiil} for exact values), 
\begin{wraptable}[8]{r}{0.6\columnwidth}
    \centering
    \resizebox{0.6\columnwidth}{!}{%
\begin{tabular}{ccccc>{\columncolor{magenta!10}}c}
\hline
Method & ERM & ARL  
& JTT 
& EIIL 
& \name \\
\hline
Train Acc & $9 2 . 7_{\pm 0 . 5}$ & $72.1_{\pm 3.6}$ & $80.2_{\pm1.7}$ & $69.7_{\pm 1.6}$ & $76.2_{\pm 2.2}$ \\
OOD Test Acc & $31.1_{\pm 4.4}$ & $61.3_{\pm 1.7}$ & $71.8_{\pm 5.3}$ & $\mathbf{7 8 . 8}_{\pm 1 . 4}$ & $\mathbf{78.8}_{\pm 0.7}$ \\
\hline
$\delta_{gap}$ & $-61.6$ & $-10.8$ & $-8.4$ & $9.1$ & $2.6$ \\
\hline
\end{tabular}
}
\caption{Comparison of average train and OOD test accuracies and their difference ($\delta_{gap}$) on Adult-Confounded dataset.}
    \label{uci}
\end{wraptable}these values are increased to $0.94$, while they are set to $0.06$ for the remaining two subgroups, to generate the Adult-Confounded dataset. In the test set, these are reversed, so that it serves as a worst-case audit to ensure that the model is not relying on subgroup membership alone in its predictions.
  
Table \ref{uci} shows that compared to other methods \name achieves superior OOD test performance with the least gap between train and test performance, indicating its low reliance on sensitive subgroup information.
\subsection{Subgroup Robustness}\label{sec:subgroup_robustness}
We evaluate our approach on four benchmark classification tasks for robustness to spurious correlations, namely on Waterbirds, CelebA, MultiNLI and CivilComments-WILDS datasets (Table \ref{tab1}).  
We follow the setup in \citet{gdro} for the first three and \citet{wilds} for CivilComments-WILDS.
\vspace{-1mm}
\paragraph{Waterbirds.} Waterbirds is a synthetic dataset created by \citet{gdro} consisting of bird images over backgrounds. The task is to classify whether a bird is a \textit{landbird} or a \textit{waterbird}. The background of the image \textit{land background} or \textit{water background}, acts a spurious correlation.
\paragraph{CelebA.} CelebA is a synthetic dataset created by \citet{celeba} containing images of celebrity faces. We classify the hair color as \textit{blonde} or \textit{not blonde}, which is spuriously correlated with the gender of the celebrity \textit{male} or \textit{female}, as done in \citet{gdro, jtt}.
\vspace{-1mm}
\paragraph{MultiNLI.} MultiNLI \citep{multinli} is a dataset of sentence pairs consisting of three classes: entailment, neutral, contradiction. Pairs are labeled based on whether the second sentence entails, is neutral with, or contradicts the first sentence, which is correlated with the presence of negation words in the second sentence \citep{gdro, jtt}.
\vspace{-1mm}
\paragraph{CivilComments-WILDS.} CivilComments-WILDS is a dataset of online comments proposed by \citet{civcoms}. The goal is to classify whether a comment is \textit{toxic} or \textit{non-toxic}, which is spuriously correlated with the mention of one or more of the following demographic attributes: male, female, White, Black, LGBTQ, Muslim, Christian, and other religion \citep{park,dixon}. Similar to previous work \citep{wilds,jtt}, we evaluate over 16 overlapping groups, one for each potential label-demographic pair. 

\paragraph{}
Although \name does not require group knowledge for training, following \citet{jtt}, we use a validation set with group labels for model selection. Table \ref{main-results} shows the average and worst-group accuracies for \name and comparison with other methods \citep{cvardro, ltf, jtt} which do not use group information. GDRO \citep{gdro}, which uses group information, acts as a benchmark. We see that on three of these datasets, \name competes with state-of-the-art algorithms that improve subgroup robustness. Interestingly, \name seems paricularly effective on the two language-based datasets. We also note that \name is not very effective on CelebA images. We believe that this is because both the spurious feature (gender) and the invariant feature (hair color) for CelebA are of similar complexity. In Section \ref{sec:celeba}, we explore this further with a similar experiment as in Section \ref{sec:define_simple} for CelebA.

\begin{table}[h!]
\centering
\resizebox{\columnwidth}{!}{%
\begin{tabular}{cccccccccc}
\hline \multirow{2}{*}{ Method }
& \multirow{2}{*}{$\begin{array}{l}\text { Group } \\
\text { labels }\end{array}$} & \multicolumn{2}{c}{ Waterbirds } & \multicolumn{2}{c}{ CelebA } & \multicolumn{2}{c}{ MultiNLI } & \multicolumn{2}{c}{ CivilComments-WILDS} \\
 & & Average & Worst-group  & Average & Worst-group & Average & Worst-group & Average & Worst-group  \\

\hline ERM & No & $97.3  $ & $72.6  $ & $95.6  $ & $47.2  $ & $82.4  $ & $67.9  $ & $92.6  $ & $57.4  $ \\
 CVaRDRO 
 & No & $96.0  $ & $75.9  $ & $82.5  $ & $64.4  $ & $82.0  $ & $68.0  $ & $92.5  $ & $60.5  $ \\
 LfF 
 & No & $91.2  $ & $78.0  $ & $85.1  $ & $\underline{77.2}  $ & $80.8  $ & $70.2  $ & $92.5  $ & $58.8  $ \\
 JTT 
 & No & $93.3  $ & $\mathbf{86.7}  $ & $88.0  $ & $\mathbf{81.1}$  & $78.6  $ & $\mathbf{72.6}  $ & $91.1  $ & $\underline{69.3}  $ \\
 \rowcolor{magenta!10}
\name  & No & $88.6$ & $\underline{84.3}$ & $84.5$ & $75.3$ & $81.4$ & $\underline{71.5}$ &  $84.2  $&  $\mathbf{74.8}  $\\
\hline GDRO 
& Yes & $93.5  $ & $91.4  $ & $92.9  $ & $88.9  $ & $81.4  $ & $77.7  $ & $88.9  $ & $69.9  $ \\
\hline
\end{tabular}%
}\caption{Average and worst-group test accuracies on benchmark datasets for subgroup robustness. \textbf{Bold} and \underline{underlined} numbers indicate the best and second-best worst-group accuracy among the methods that don't use group labels.} 
 \label{main-results}
\end{table}
\vspace{-1mm}
\subsection{Fairness Application: Bias in Occupation Prediction}
The Bios dataset
\citep{bios-data,snob} is a large-scale dataset of more than 300k biographies scrapped from the internet. The goal is to predict a person's occupation based on their bio. Based on this task, \citet{snob} formalizes a notion of social norm bias (SNoB). SNoB captures the extent to which predictions align with gender norms associated with specific occupations. In addition to gender-specific pronouns, these norms encompass other characteristics mentioned in the bios. They represent implicit expectations of how specific groups are expected to behave. \citet{snob} characterizes SNoB as a form of algorithmic unfairness arising from the associations between an algorithm's predictions and individuals' adherence to inferred social norms. They also show that adherence to or deviations from social norms can result in harm in many contexts and that SNoB can persist even after the application of some fairness interventions. We note that SNoB is a distinct type of bias, and existing de-biasing methods have not been evaluated on this task.

\begin{wraptable}[8]{r}{0.35\columnwidth}
    \centering
    \vspace{-3mm}
    \resizebox{0.35\columnwidth}{!}{%
\begin{tabular}{ccc>{\columncolor{magenta!10}}c}
\hline Method & ERM & Decoupled & \name \\
\hline
Accuracy & $0.95$ & $0.94$ & $\mathbf{0.96}$\\ 
$\rho(p_c,r_c)$ & $ 0.66$ & $ 0.60$ & $ \mathbf{0.38}$ \\  
\hline
\end{tabular}
}\caption{Comparison of accuracy in occupation prediction and the correlation between gender and occupation prediction on Bios data.} 
\label{bios}
\end{wraptable}
To quantify SNoB, the authors utilize the Spearman rank correlation coefficient $\rho(p_c, r_c)$, where $p_c$ represents the fraction of bios associated with occupation $c$ that mention the pronoun `she', and $r_c$ measures the correlation between occupation predictions and gender predictions. The authors employ separate one-vs-all classifiers for each occupation and obtain the occupation prediction for a given bio using these classifiers. For gender predictions, they train occupation-specific models to determine the gender-based group membership (female or not) based on a person's bio, and use the predictions from these models. A higher value of $\rho(p_c, r_c)$ represents a larger social norm bias, which indicates that in male-dominated occupations, the algorithm achieves higher accuracy on bios that align with inferred masculine norms, and vice-versa. Table \ref{bios} shows results on the Bios data. We compare with a group fairness approach, Decoupled \citep{decoupled} that trains separate models for each gender, in order to mitigate gender bias. We see that  \name address SNoB bias better than ERM and Decoupled, achieving a lower $\rho(p_c,r_c)$ and improved accuracy.
\section{Additional Experiments}

This section presents additional experiments, including a comparison between the original and estimated CMI, and an evaluation of the complexity of the invariant and surrogate features in the CelebA dataset.
\subsection{Evaluating the Estimated CMI}
\label{A1}
\begin{wraptable}[7]{r}{0.37\columnwidth}
    \centering
    \vspace{-3mm}
    \resizebox{0.34\columnwidth}{!}{%
\begin{tabular}{ccc}
\hline Method 
& 10 classes  & 200 classes \\
\hline $I(M,M_s|Y)$  
& $14.1\pm 3.3$ & $23.3\pm2.7$\\ 
$\hat{I}_n(M,M_s|Y)$  
& $20.17\pm4.2$ & $24.3\pm3.3$

\\ 
\hline
\end{tabular}
}\caption{Comparison of computation times (in milliseconds) for the original CMI and the estimated CMI in \eqref{cmi-est-eq}.} 
\label{comp_time}
\end{wraptable}

In this section, we evaluate the reliability and scalability of the CMI estimate in \eqref{cmi-est-eq} compared to the original CMI, which is computed with discretized model outputs. We consider CMNIST data with the hyperparameter values as mentioned in Appendix \ref{cmnist-hp} for the results in this section.

Table \ref{comp_time} compares the times (in milliseconds) to compute the original CMI and the estimated CMI, using batch size $64$, for 10 classes and 200 classes. We see that the computation time for the estimated CMI does not increase significantly as the number of classes increases.  

\begin{figure}[h!]
  \centering
  \begin{minipage}[b]{0.61\textwidth}
    \centering
    \includegraphics[height=90pt]{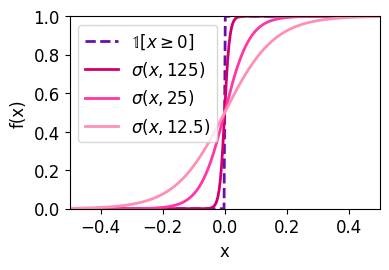} \includegraphics[height=90pt]{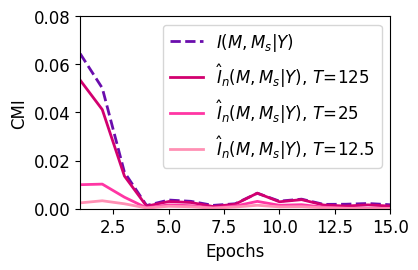}  
    \caption{(Left) Sigmoid functions for different values of the temperature parameter $T$. (Right) Comparison of the original CMI with the estimated CMI using different values of $T$. 
    }
    \label{fig:cmi-time2}
  \end{minipage}
  \hfill
  \begin{minipage}[b]{0.37\textwidth}
    \centering
    \includegraphics[height=85pt]{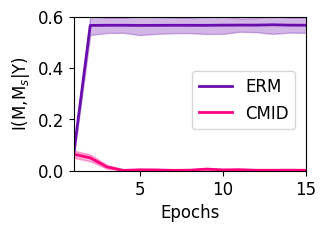}  
    \caption{Comparison of the original CMI for models trained with ERM and CMID as a function of training time.}
    \label{fig:cmi-time}
  \end{minipage}
  
\end{figure}
Fig. \ref{fig:cmi-time2} compares the estimated CMI for different values of the temperature parameter $T$ with the original CMI. The values shown in the figure are computed with batch size $1000$, while for training, we use the estimated CMI with $T=12.5$ and batch size $64$. We see that as $T$ increases, the sigmoid approximates the indicator function more closely and the estimated CMI tends closer to the original CMI. 

Fig. \ref{fig:cmi-time} compares the CMI computed with discretized outputs using batch size $1000$, when the model is trained with ERM or CMID. We see that regularizing the estimated CMI causes the original CMI to decrease over time, whereas training with ERM leads to an increase in the CMI. This shows that the solution learned by ERM makes predictions aligned with those of the simple model, whereas our approach CMID learns a model which attains low CMI with the simple model. 

\subsection{CelebA: Invariant \& Spurious Features have Similar Complexity}
\label{sec:celeba}

\begin{wraptable}[8]{r}{0.5\columnwidth}
\vspace{-1mm}
\centering
\resizebox{0.5\columnwidth}{!}{%
\begin{tabular}{lcccc}    
\hline  \multicolumn{2}{c}{ Predict invariant feature } & \multicolumn{2}{c}{ Predict surrogate feature} \\
    Train & Test   & Train & Test  \\
\hline 
 $89.1\pm 1.5$ &
$84.3\pm 0.6$ & $92.4\pm 2.4$ & $88.3\pm1.6$ \\

\hline
\end{tabular}}
\caption{Comparison between performance for predicting the simple feature and the complex feature on CelebA dataset.}
    \label{celeb}
\end{wraptable}
In our subgroup robustness experiment for CelebA (Table \ref{main-results} in Section \ref{sec:subgroup_robustness}), we found that our method did not yield a significant improvement in worst-group accuracy. We investigate this further in this section. We show that for the CelebA dataset, the complexity of the invariant and surrogate features are actually quite  similar. The experiment is similar to the experiments we did for CMNIST and Waterbirds in Section \ref{sec:define_simple}. We create a subset of the CelebA dataset by sampling an equal number of samples from all four subgroups. 

Table \ref{celeb} presents a comparison of the results when predicting the invariant feature (hair color) and the surrogate feature (gender) using the simple model (\ts{2DConvNet2}). We observe that the performance for both tasks is comparable, suggesting that the features exhibit similar complexity. We contrast these results with those for CMNIST and Waterbirds in Table \ref{simple-complex}. For CMNIST and Waterbirds, there was a significant difference in the accuracy to which the simple model could predict the invariant and surrogate feature. For CelebA, the difference is much smaller which suggests that spurious features are \emph{not} simpler than  invariant features for this dataset---explaining why our method is not as effective for it.
\section{Discussion and Conclusion}
\label{sec:discussion}
We proposed a new framework (CMID) to mitigate simplicity bias, and showed that it yields improvements over ERM and many other previous approaches across a number of OOD generalization, robustness and fairness benchmarks. We conclude with some discussion on differences with prior work, limitations of our approach, and directions for future work.

\paragraph{Connections with Prior Work.}
In general, prior work focuses on one or two of the applications that we consider, while our approach proves effective across several datasets for all cases. 

Several methods that aim to improve subgroup robustness or OOD generalization, such as IRM \citep{irm}, GDRO \citep{gdro}, PI \citep{pi} and BLOOD \citep{blood}, require knowledge of group or environment labels. This additional information is not always available and consequently these methods have limited applicability in such cases. Approaches like JTT \citep{jtt} and EIIL \citep{eiil} overcome this requirement. However, these methods rely on incorrect predictions \citep{jtt,eiil} of one model to train another model by upweighting such samples \citep{jtt} 
or learning features that generalize across such samples \citep{eiil}. As a result, they may not be effective in settings where the spurious feature is highly predictive. Such settings can be relevant when our goal is to mitigate simplicity bias. For example, in the slab dataset (Figure \ref{fig:slab}), the simple model achieves perfect accuracy, but has a much worse margin compared to the global max margin predictor. Our approach is still effective in such cases because it relies on prediction probabilities rather than thresholded predictions, which enables the differentiation between features that result in smaller margin or low confidence predictions and those that lead to better margin or high confidence predictions.

We note that debiasing/feature diversification methods \citep{ltf,debias,debian,bonsai,evading-sb} can often be computationally expensive. Some of these involve training two complex models simultaneously \citep{ltf,debias} or alternatively \citep{debian}. \citet{bonsai,evading-sb} involve training multiple models and returning the average or selecting the best one among them, respectively. In contrast, our results show that the simple approach to train a single model with the CMI regularization can prove effective across several settings. While methods like \citet{agdisag,divdis} are simpler, they require access to additional data from a target domain to encourage disagreement between two models for feature diversification. Our approach does not have such a requirement and allows for diversification directly on the training data.

In addition, our approach is theoretically grounded and explicit in its assumptions, which could make it easier for a practitioner to evaluate and use. In particular, our definition of spurious features gives an explicit characterization of features/biases that we regard as undesirable and our approach seeks to reduce, which can allow the user to understand the situations when the approach might be effective. We note that most approaches developed to improve subgroup robustness or OOD generalization either lack theoretical guarantees \citep{pi,blood,jtt} or show results for either subgroup robustness or OOD generalization, but not both. \citet{gdro,cvardro} seek to optimize the worst-case risk over a small set of groups and hence, guarantees on worst-group risk developed in prior work on DRO which considers all subsets of the data as the groups of interest, apply in these cases. However, these methods may not generalize to unseen environments, as is often times the case in the OOD generalization problem. For instance, in our experiments on the Camelyon17 dataset, we observe that GDRO \citep{gdro} is not very effective in terms of performance on the test set, which consists of samples from unseen hospitals (environments). On the other hand, \citet{irm} prove that IRM can recover the optimal invariant predictor that can generalize to any environment, under certain assumptions on the training data. However, follow up work \citep{rosenfeld} shows that IRM may fail to recover such a predictor if the training set does not contain samples from several environments. In contrast, we show that our approach reduces reliance on the spurious feature in a Gaussian mixture based data setting, and also obtain an OOD generalization guarantee in a causal learning framework, by establishing a condition on the data under which the invariant predictor can be recovered without knowledge of the environment labels.

\paragraph{Limitations.} 

We note that like any other regularization or inductive bias, CMID may not be effective for \emph{every} task. For certain tasks, spurious features as defined by us in Section \ref{sec:define_simple} may not actually be spurious, and we may not always want to reduce reliance on features that are simple and highly predictive of the label. However, it is worth noting here that it is impossible for a single algorithm to perform well in all cases. This is established by the No Free Lunch theorem: there is no inductive bias that is suitable for all tasks \citep{nfl2} and relatedly, it is impossible to generalize without certain assumptions on the train and test distributions \citep{nfl1}.

In most cases, we assume that the test data is i.i.d. and indeed, simplicity bias is useful in explaining in-distribution generalization of NNs in these cases. However, these observations may not necessarily extend to situations where the i.i.d. assumption on the test set is violated. Generalization under distribution shift is significantly different and in such cases, simplicity bias can prevent the model from learning more complex, task-relevant features that may generalize better. Consequently, alleviating simplicity bias can be a useful inductive bias in such cases. This is consistent with our experimental evaluations on several datasets from various modalities and domains. 

 Even in situations where mitigating simplicity bias is useful, there can be cases where the separation between the feature complexity of spurious and invariant features may not be very large. This can make the selection of an appropriate simple model class for our approach challenging. Indeed, in our experiments, we found that for the CelebA dataset, the spurious and task-relevant features are not significantly different in terms of complexity (Section \ref{sec:celeba}). Consequently, our approach does not lead to much improvement in the worst-group accuracy on this dataset. 

We also note that in general, use of methods designed to improve OOD generalization and subgroup robustness can lead to a drop in the accuracy on the training set or the i.i.d. test set, compared to ERM. This is seen in some of our experimental results as well. This is because a model trained with the ERM objective relies strongly on the spurious feature(s), which is predictive on the train set as well as the i.i.d. test set. In contrast, improvement on OOD test sets is achieved by leveraging other features, which may be less predictive on the train set but generalize better. 

\paragraph{Future Work.} A natural direction of future work is to further explore the capabilities and limitations of our approach, and to also further understand its theoretical properties. More broadly, our work suggests auditing large models with respect to much simpler models can lead to improved properties of the larger models along certain robustness and fairness axes. It would be interesting to explore the power of similar approaches for other desiderata, and to understand its capabilities for fine-tuning large pre-trained models.

\section*{Acknowledgements}
This work was supported by NSF CAREER Award CCF-2239265 and a Amazon Research Award. The authors acknowledge use of the Amazon Elastic Compute Cloud and USC Advanced Research Computing's Discovery cluster. BV thanks Puneesh Deora for helpful discussions and feedback, and Deqing Fu for discussion on simplicity bias.
\bibliography{refs}

\begin{thebibliography}{66}
\providecommand{\natexlab}[1]{#1}
\providecommand{\url}[1]{\texttt{#1}}
\expandafter\ifx\csname urlstyle\endcsname\relax
  \providecommand{\doi}[1]{doi: #1}\else
  \providecommand{\doi}{doi: \begingroup \urlstyle{rm}\Url}\fi

\bibitem[Arjovsky et~al.(2020)Arjovsky, Bottou, Gulrajani, and Lopez-Paz]{irm}
Martin Arjovsky, Léon Bottou, Ishaan Gulrajani, and David Lopez-Paz.
\newblock Invariant risk minimization, 2020.

\bibitem[Arpit et~al.(2017)Arpit, Jastrz{\k{e}}bski, Ballas, Krueger, Bengio,
  Kanwal, Maharaj, Fischer, Courville, Bengio, et~al.]{arpit2017closer}
Devansh Arpit, Stanis{\l}aw Jastrz{\k{e}}bski, Nicolas Ballas, David Krueger,
  Emmanuel Bengio, Maxinder~S Kanwal, Tegan Maharaj, Asja Fischer, Aaron
  Courville, Yoshua Bengio, et~al.
\newblock A closer look at memorization in deep networks.
\newblock In \emph{International conference on machine learning}, pages
  233--242. PMLR, 2017.

\bibitem[Bae et~al.(2022)Bae, Choi, and Lee]{blood}
Jun-Hyun Bae, Inchul Choi, and Minho Lee.
\newblock {BLOOD}: Bi-level learning framework for out-of-distribution
  generalization, 2022.
\newblock URL \url{https://openreview.net/forum?id=Cm08egNmrl3}.

\bibitem[Bahng et~al.(2020)Bahng, Chun, Yun, Choo, and Oh]{debias}
Hyojin Bahng, Sanghyuk Chun, Sangdoo Yun, Jaegul Choo, and Seong~Joon Oh.
\newblock Learning de-biased representations with biased representations.
\newblock In Hal~Daumé III and Aarti Singh, editors, \emph{Proceedings of the
  37th International Conference on Machine Learning}, volume 119 of
  \emph{Proceedings of Machine Learning Research}, pages 528--539. PMLR, 13--18
  Jul 2020.
\newblock URL \url{https://proceedings.mlr.press/v119/bahng20a.html}.

\bibitem[Bao et~al.(2021)Bao, Chang, and Barzilay]{pi}
Yujia Bao, Shiyu Chang, and Regina Barzilay.
\newblock Predict then interpolate: A simple algorithm to learn stable
  classifiers.
\newblock In \emph{International Conference on Machine Learning}. PMLR, 2021.

\bibitem[Borkan et~al.(2019)Borkan, Dixon, Sorensen, Thain, and
  Vasserman]{civcoms}
Daniel Borkan, Lucas Dixon, Jeffrey Sorensen, Nithum Thain, and Lucy Vasserman.
\newblock Nuanced metrics for measuring unintended bias with real data for text
  classification.
\newblock In \emph{Companion Proceedings of The 2019 World Wide Web
  Conference}, 2019.

\bibitem[Bándi et~al.(2019)Bándi, Geessink, Manson, Van~Dijk, Balkenhol,
  Hermsen, Ehteshami~Bejnordi, Lee, Paeng, Zhong, Li, Zanjani, Zinger, Fukuta,
  Komura, Ovtcharov, Cheng, Zeng, Thagaard, Dahl, Lin, Chen, Jacobsson,
  Hedlund, Çetin, Halıcı, Jackson, Chen, Both, Franke, Küsters-Vandevelde,
  Vreuls, Bult, van Ginneken, van~der Laak, and Litjens]{camelyon}
Péter Bándi, Oscar Geessink, Quirine Manson, Marcory Van~Dijk, Maschenka
  Balkenhol, Meyke Hermsen, Babak Ehteshami~Bejnordi, Byungjae Lee, Kyunghyun
  Paeng, Aoxiao Zhong, Quanzheng Li, Farhad~Ghazvinian Zanjani, Svitlana
  Zinger, Keisuke Fukuta, Daisuke Komura, Vlado Ovtcharov, Shenghua Cheng,
  Shaoqun Zeng, Jeppe Thagaard, Anders~B. Dahl, Huangjing Lin, Hao Chen, Ludwig
  Jacobsson, Martin Hedlund, Melih Çetin, Eren Halıcı, Hunter Jackson,
  Richard Chen, Fabian Both, Jörg Franke, Heidi Küsters-Vandevelde, Willem
  Vreuls, Peter Bult, Bram van Ginneken, Jeroen van~der Laak, and Geert
  Litjens.
\newblock From detection of individual metastases to classification of lymph
  node status at the patient level: The camelyon17 challenge.
\newblock \emph{IEEE Transactions on Medical Imaging}, 38\penalty0
  (2):\penalty0 550--560, 2019.
\newblock \doi{10.1109/TMI.2018.2867350}.

\bibitem[Chang et~al.(2020)Chang, Zhang, Yu, and Jaakkola]{chang-ir}
Shiyu Chang, Yang Zhang, Mo~Yu, and Tommi Jaakkola.
\newblock Invariant rationalization.
\newblock In Hal~Daumé III and Aarti Singh, editors, \emph{Proceedings of the
  37th International Conference on Machine Learning}, volume 119 of
  \emph{Proceedings of Machine Learning Research}, pages 1448--1458. PMLR,
  13--18 Jul 2020.
\newblock URL \url{https://proceedings.mlr.press/v119/chang20c.html}.

\bibitem[Cheng et~al.(2023)Cheng, De-Arteaga, Mackey, and Kalai]{snob}
Myra Cheng, Maria De-Arteaga, Lester Mackey, and Adam~Tauman Kalai.
\newblock Social norm bias: residual harms of fairness-aware algorithms.
\newblock \emph{Data Mining and Knowledge Discovery}, pages 1--27, 2023.

\bibitem[Creager et~al.(2021)Creager, Jacobsen, and Zemel]{eiil}
Elliot Creager, Joern-Henrik Jacobsen, and Richard Zemel.
\newblock Environment inference for invariant learning.
\newblock In Marina Meila and Tong Zhang, editors, \emph{Proceedings of the
  38th International Conference on Machine Learning}, volume 139 of
  \emph{Proceedings of Machine Learning Research}, pages 2189--2200. PMLR,
  18--24 Jul 2021.
\newblock URL \url{https://proceedings.mlr.press/v139/creager21a.html}.

\bibitem[Dagaev et~al.(2023)Dagaev, Roads, Luo, Barry, Patil, and
  Love]{toogood}
Nikolay Dagaev, Brett~D. Roads, Xiaoliang Luo, Daniel~N. Barry, Kaustubh~R.
  Patil, and Bradley~C. Love.
\newblock A too-good-to-be-true prior to reduce shortcut reliance.
\newblock \emph{Pattern Recogn. Lett.}, 166\penalty0 (C):\penalty0 164–171,
  feb 2023.
\newblock ISSN 0167-8655.
\newblock \doi{10.1016/j.patrec.2022.12.010}.
\newblock URL \url{https://doi.org/10.1016/j.patrec.2022.12.010}.

\bibitem[De-Arteaga et~al.(2019)De-Arteaga, Romanov, Wallach, Chayes, Borgs,
  Chouldechova, Geyik, Kenthapadi, and Kalai]{bios-data}
Maria De-Arteaga, Alexey Romanov, Hanna Wallach, Jennifer Chayes, Christian
  Borgs, Alexandra Chouldechova, Sahin Geyik, Krishnaram Kenthapadi, and
  Adam~Tauman Kalai.
\newblock Bias in bios: A case study of semantic representation bias in a
  high-stakes setting.
\newblock In \emph{Proceedings of the Conference on Fairness, Accountability,
  and Transparency}, FAT* '19, page 120–128, New York, NY, USA, 2019.
  Association for Computing Machinery.
\newblock ISBN 9781450361255.
\newblock \doi{10.1145/3287560.3287572}.
\newblock URL \url{https://doi.org/10.1145/3287560.3287572}.

\bibitem[Deng(2012)]{mnist-cite}
Li~Deng.
\newblock The mnist database of handwritten digit images for machine learning
  research.
\newblock \emph{IEEE Signal Processing Magazine}, 29\penalty0 (6):\penalty0
  141--142, 2012.

\bibitem[Devlin et~al.(2019)Devlin, Chang, Lee, and Toutanova]{devlin-bert}
Jacob Devlin, Ming-Wei Chang, Kenton Lee, and Kristina Toutanova.
\newblock {BERT}: Pre-training of deep bidirectional transformers for language
  understanding.
\newblock In \emph{Proceedings of the 2019 Conference of the North {A}merican
  Chapter of the Association for Computational Linguistics: Human Language
  Technologies, Volume 1 (Long and Short Papers)}, pages 4171--4186,
  Minneapolis, Minnesota, June 2019. Association for Computational Linguistics.
\newblock \doi{10.18653/v1/N19-1423}.
\newblock URL \url{https://aclanthology.org/N19-1423}.

\bibitem[Dixon et~al.(2018)Dixon, Li, Sorensen, Thain, and Vasserman]{dixon}
Lucas Dixon, John Li, Jeffrey Sorensen, Nithum Thain, and Lucy Vasserman.
\newblock Measuring and mitigating unintended bias in text classification.
\newblock 2018.

\bibitem[Duchi et~al.(2019)Duchi, Hashimoto, and Namkoong]{cvardro}
John~C. Duchi, Tatsunori Hashimoto, and Hongseok Namkoong.
\newblock Distributionally robust losses against mixture covariate shifts,
  2019.

\bibitem[Dwork et~al.(2012)Dwork, Hardt, Pitassi, Reingold, and Zemel]{dem-par}
Cynthia Dwork, Moritz Hardt, Toniann Pitassi, Omer Reingold, and Richard Zemel.
\newblock Fairness through awareness.
\newblock In \emph{Proceedings of the 3rd Innovations in Theoretical Computer
  Science Conference}, ITCS '12, page 214–226, New York, NY, USA, 2012.
  Association for Computing Machinery.
\newblock ISBN 9781450311151.
\newblock \doi{10.1145/2090236.2090255}.
\newblock URL \url{https://doi.org/10.1145/2090236.2090255}.

\bibitem[Dwork et~al.(2018)Dwork, Immorlica, Kalai, and Leiserson]{decoupled}
Cynthia Dwork, Nicole Immorlica, Adam~Tauman Kalai, and Max Leiserson.
\newblock Decoupled classifiers for group-fair and efficient machine learning.
\newblock In Sorelle~A. Friedler and Christo Wilson, editors, \emph{Proceedings
  of the 1st Conference on Fairness, Accountability and Transparency},
  volume~81 of \emph{Proceedings of Machine Learning Research}, pages 119--133.
  PMLR, 23--24 Feb 2018.
\newblock URL \url{https://proceedings.mlr.press/v81/dwork18a.html}.

\bibitem[Frankle and Carbin(2019)]{lottery}
Jonathan Frankle and Michael Carbin.
\newblock The lottery ticket hypothesis: Finding sparse, trainable neural
  networks.
\newblock In \emph{International Conference on Learning Representations}, 2019.
\newblock URL \url{https://openreview.net/forum?id=rJl-b3RcF7}.

\bibitem[Gao et~al.(2022)Gao, Sagawa, Koh, Hashimoto, and Liang]{gao-ood}
Irena Gao, Shiori Sagawa, Pang~Wei Koh, Tatsunori Hashimoto, and Percy Liang.
\newblock Out-of-distribution robustness via targeted augmentations.
\newblock In \emph{NeurIPS 2022 Workshop on Distribution Shifts: Connecting
  Methods and Applications}, 2022.
\newblock URL \url{https://openreview.net/forum?id=Bcg0It4i1g}.

\bibitem[Geirhos et~al.(2020)Geirhos, Jacobsen, Michaelis, Zemel, Brendel,
  Bethge, and Wichmann]{geirhos2020shortcut}
Robert Geirhos, J{\"o}rn-Henrik Jacobsen, Claudio Michaelis, Richard Zemel,
  Wieland Brendel, Matthias Bethge, and Felix~A Wichmann.
\newblock Shortcut learning in deep neural networks.
\newblock \emph{Nature Machine Intelligence}, 2\penalty0 (11):\penalty0
  665--673, 2020.

\bibitem[Geirhos et~al.(2022)Geirhos, Rubisch, Michaelis, Bethge, Wichmann, and
  Brendel]{geirhos}
Robert Geirhos, Patricia Rubisch, Claudio Michaelis, Matthias Bethge, Felix~A.
  Wichmann, and Wieland Brendel.
\newblock Imagenet-trained cnns are biased towards texture; increasing shape
  bias improves accuracy and robustness, 2022.

\bibitem[Gururangan et~al.(2018)Gururangan, Swayamdipta, Levy, Schwartz,
  Bowman, and Smith]{guru-nlp}
Suchin Gururangan, Swabha Swayamdipta, Omer Levy, Roy Schwartz, Samuel Bowman,
  and Noah~A. Smith.
\newblock Annotation artifacts in natural language inference data.
\newblock In \emph{Proceedings of the 2018 Conference of the North {A}merican
  Chapter of the Association for Computational Linguistics: Human Language
  Technologies, Volume 2 (Short Papers)}, pages 107--112, New Orleans,
  Louisiana, June 2018. Association for Computational Linguistics.
\newblock \doi{10.18653/v1/N18-2017}.
\newblock URL \url{https://aclanthology.org/N18-2017}.

\bibitem[Hardt et~al.(2016)Hardt, Price, and Srebro]{eq-opp}
Moritz Hardt, Eric Price, and Nathan Srebro.
\newblock Equality of opportunity in supervised learning.
\newblock In \emph{Proceedings of the 30th International Conference on Neural
  Information Processing Systems}, NIPS'16, page 3323–3331, Red Hook, NY,
  USA, 2016. Curran Associates Inc.
\newblock ISBN 9781510838819.

\bibitem[Hashimoto et~al.(2018)Hashimoto, Srivastava, Namkoong, and
  Liang]{replossmin}
Tatsunori~B. Hashimoto, Megha Srivastava, Hongseok Namkoong, and Percy Liang.
\newblock Fairness without demographics in repeated loss minimization.
\newblock In \emph{International Conference on Machine Learning}, 2018.

\bibitem[He et~al.(2016)He, Zhang, Ren, and Sun]{resnet50}
Kaiming He, Xiangyu Zhang, Shaoqing Ren, and Jian Sun.
\newblock Deep residual learning for image recognition.
\newblock In \emph{2016 IEEE Conference on Computer Vision and Pattern
  Recognition (CVPR)}, pages 770--778, 2016.
\newblock \doi{10.1109/CVPR.2016.90}.

\bibitem[Hebert-Johnson et~al.(2018)Hebert-Johnson, Kim, Reingold, and
  Rothblum]{multicalib}
Ursula Hebert-Johnson, Michael Kim, Omer Reingold, and Guy Rothblum.
\newblock Multicalibration: Calibration for the
  ({C}omputationally-identifiable) masses.
\newblock In Jennifer Dy and Andreas Krause, editors, \emph{Proceedings of the
  35th International Conference on Machine Learning}, volume~80 of
  \emph{Proceedings of Machine Learning Research}, pages 1939--1948. PMLR,
  10--15 Jul 2018.
\newblock URL \url{https://proceedings.mlr.press/v80/hebert-johnson18a.html}.

\bibitem[Jia et~al.(2020)Jia, Meng, Zhao, and Chang]{zhao-nlp}
Shengyu Jia, Tao Meng, Jieyu Zhao, and Kai-Wei Chang.
\newblock Mitigating gender bias amplification in distribution by posterior
  regularization.
\newblock In \emph{Proceedings of the 58th Annual Meeting of the Association
  for Computational Linguistics}, pages 2936--2942, Online, July 2020.
  Association for Computational Linguistics.
\newblock \doi{10.18653/v1/2020.acl-main.264}.
\newblock URL \url{https://aclanthology.org/2020.acl-main.264}.

\bibitem[Kamiran and Calders(2012)]{rewt}
Faisal Kamiran and Toon Calders.
\newblock Data preprocessing techniques for classification without
  discrimination.
\newblock \emph{Knowl. Inf. Syst.}, 33\penalty0 (1):\penalty0 1–33, oct 2012.
\newblock ISSN 0219-1377.
\newblock \doi{10.1007/s10115-011-0463-8}.
\newblock URL \url{https://doi.org/10.1007/s10115-011-0463-8}.

\bibitem[Kirichenko et~al.(2022)Kirichenko, Izmailov, and Wilson]{lastlayer}
Polina Kirichenko, Pavel Izmailov, and Andrew~Gordon Wilson.
\newblock Last layer re-training is sufficient for robustness to spurious
  correlations.
\newblock In \emph{International Conference on Machine Learning}, 2022.

\bibitem[Koh et~al.(2021)Koh, Sagawa, Marklund, Xie, Zhang, Balsubramani, Hu,
  Yasunaga, Phillips, Gao, et~al.]{wilds}
Pang~Wei Koh, Shiori Sagawa, Henrik Marklund, Sang~Michael Xie, Marvin Zhang,
  Akshay Balsubramani, Weihua Hu, Michihiro Yasunaga, Richard~Lanas Phillips,
  Irena Gao, et~al.
\newblock Wilds: A benchmark of in-the-wild distribution shifts.
\newblock In \emph{International Conference on Machine Learning}, pages
  5637--5664. PMLR, 2021.

\bibitem[Lahoti et~al.(2020)Lahoti, Beutel, Chen, Lee, Prost, Thain, Wang, and
  Chi]{lahoti}
Preethi Lahoti, Alex Beutel, Jilin Chen, Kang Lee, Flavien Prost, Nithum Thain,
  Xuezhi Wang, and Ed~H. Chi.
\newblock Fairness without demographics through adversarially reweighted
  learning.
\newblock In \emph{Proceedings of the 34th International Conference on Neural
  Information Processing Systems}, NIPS'20, Red Hook, NY, USA, 2020. Curran
  Associates Inc.
\newblock ISBN 9781713829546.

\bibitem[Lee et~al.(2022)Lee, Yao, and Finn]{divdis}
Yoonho Lee, Huaxiu Yao, and Chelsea Finn.
\newblock Diversify and disambiguate: Learning from underspecified data.
\newblock In \emph{ICML 2022: Workshop on Spurious Correlations, Invariance and
  Stability}, 2022.
\newblock URL \url{https://openreview.net/forum?id=lceGyKleset}.

\bibitem[Leisch and Dimitriadou(2021)]{uci2}
Friedrich Leisch and Evgenia Dimitriadou.
\newblock \emph{mlbench: Machine Learning Benchmark Problems}, 2021.
\newblock R package version 2.1-3.1.

\bibitem[Li et~al.(2022)Li, Hoogs, and Xu]{debian}
Zhiheng Li, Anthony Hoogs, and Chenliang Xu.
\newblock Discover and mitigate unknown biases with debiasing alternate
  networks.
\newblock In \emph{Computer Vision – ECCV 2022: 17th European Conference, Tel
  Aviv, Israel, October 23–27, 2022, Proceedings, Part XIII}, page 270–288,
  Berlin, Heidelberg, 2022. Springer-Verlag.
\newblock ISBN 978-3-031-19777-2.
\newblock \doi{10.1007/978-3-031-19778-9_16}.
\newblock URL \url{https://doi.org/10.1007/978-3-031-19778-9_16}.

\bibitem[Liu et~al.(2021{\natexlab{a}})Liu, Haghgoo, Chen, Raghunathan, Koh,
  Sagawa, Liang, and Finn]{jtt}
Evan~Z Liu, Behzad Haghgoo, Annie~S Chen, Aditi Raghunathan, Pang~Wei Koh,
  Shiori Sagawa, Percy Liang, and Chelsea Finn.
\newblock Just train twice: Improving group robustness without training group
  information.
\newblock In Marina Meila and Tong Zhang, editors, \emph{Proceedings of the
  38th International Conference on Machine Learning}, volume 139 of
  \emph{Proceedings of Machine Learning Research}, pages 6781--6792. PMLR,
  18--24 Jul 2021{\natexlab{a}}.
\newblock URL \url{https://proceedings.mlr.press/v139/liu21f.html}.

\bibitem[Liu et~al.(2021{\natexlab{b}})Liu, Hu, Cui, Li, and Shen]{hrm}
Jiashuo Liu, Zheyuan Hu, Peng Cui, B.~Li, and Zheyan Shen.
\newblock Heterogeneous risk minimization.
\newblock In \emph{International Conference on Machine Learning},
  2021{\natexlab{b}}.

\bibitem[Liu et~al.(2015)Liu, Luo, Wang, and Tang]{celeba}
Ziwei Liu, Ping Luo, Xiaogang Wang, and Xiaoou Tang.
\newblock Deep learning face attributes in the wild.
\newblock In \emph{Proceedings of International Conference on Computer Vision
  (ICCV)}, December 2015.

\bibitem[Morwani et~al.(2023)Morwani, Batra, Jain, and Netrapalli]{sb1layer}
Depen Morwani, Jatin Batra, Prateek Jain, and Praneeth Netrapalli.
\newblock Simplicity bias in 1-hidden layer neural networks, 2023.

\bibitem[Nakkiran et~al.(2019)Nakkiran, Kaplun, Kalimeris, Yang, Edelman,
  Zhang, and Barak]{inc-comp}
Preetum Nakkiran, Gal Kaplun, Dimitris Kalimeris, Tristan Yang, Benjamin~L.
  Edelman, Fred Zhang, and Boaz Barak.
\newblock \emph{SGD on Neural Networks Learns Functions of Increasing
  Complexity}.
\newblock Curran Associates Inc., Red Hook, NY, USA, 2019.

\bibitem[Nakkiran et~al.(2020)Nakkiran, Kaplun, Bansal, Yang, Barak, and
  Sutskever]{NakkiranDescent}
Preetum Nakkiran, Gal Kaplun, Yamini Bansal, Tristan Yang, Boaz Barak, and Ilya
  Sutskever.
\newblock Deep double descent: Where bigger models and more data hurt.
\newblock In \emph{International Conference on Learning Representations}, 2020.
\newblock URL \url{https://openreview.net/forum?id=B1g5sA4twr}.

\bibitem[Nam et~al.(2020)Nam, Cha, Ahn, Lee, and Shin]{ltf}
Junhyun Nam, Hyuntak Cha, Sungsoo Ahn, Jaeho Lee, and Jinwoo Shin.
\newblock Learning from failure: De-biasing classifier from biased classifier.
\newblock In H.~Larochelle, M.~Ranzato, R.~Hadsell, M.F. Balcan, and H.~Lin,
  editors, \emph{Advances in Neural Information Processing Systems}, volume~33,
  pages 20673--20684. Curran Associates, Inc., 2020.
\newblock URL
  \url{https://proceedings.neurips.cc/paper_files/paper/2020/file/eddc3427c5d77843c2253f1e799fe933-Paper.pdf}.

\bibitem[Newman et~al.(1998)Newman, Hettich, Blake, and Merz]{uci1}
D.J. Newman, S.~Hettich, C.L. Blake, and C.J. Merz.
\newblock Uci repository of machine learning databases, 1998.
\newblock URL \url{http://www.ics.uci.edu/~mlearn/MLRepository.html}.

\bibitem[Pagliardini et~al.(2023)Pagliardini, Jaggi, Fleuret, and
  Karimireddy]{agdisag}
Matteo Pagliardini, Martin Jaggi, Fran{\c{c}}ois Fleuret, and Sai~Praneeth
  Karimireddy.
\newblock Agree to disagree: Diversity through disagreement for better
  transferability.
\newblock In \emph{The Eleventh International Conference on Learning
  Representations}, 2023.
\newblock URL \url{https://openreview.net/forum?id=K7CbYQbyYhY}.

\bibitem[Park et~al.(2018)Park, Shin, and Fung]{park}
Ji~Ho Park, Jamin Shin, and Pascale Fung.
\newblock Reducing gender bias in abusive language detection, 2018.

\bibitem[Pezeshki et~al.(2021)Pezeshki, Kaba, Bengio, Courville, Precup, and
  Lajoie]{gs}
Mohammad Pezeshki, S{\'e}kou-Oumar Kaba, Yoshua Bengio, Aaron Courville, Doina
  Precup, and Guillaume Lajoie.
\newblock Gradient starvation: A learning proclivity in neural networks.
\newblock In A.~Beygelzimer, Y.~Dauphin, P.~Liang, and J.~Wortman Vaughan,
  editors, \emph{Advances in Neural Information Processing Systems}, 2021.
\newblock URL \url{https://openreview.net/forum?id=aExAsh1UHZo}.

\bibitem[Qiu et~al.(2023)Qiu, Potapczynski, Izmailov, and Wilson]{afr}
Shikai Qiu, Andres Potapczynski, Pavel Izmailov, and Andrew~Gordon Wilson.
\newblock {Simple and Fast Group Robustness by Automatic Feature Reweighting}.
\newblock \emph{International Conference on Machine Learning (ICML)}, 2023.

\bibitem[Rosenfeld et~al.(2021)Rosenfeld, Ravikumar, and Risteski]{rosenfeld}
Elan Rosenfeld, Pradeep~Kumar Ravikumar, and Andrej Risteski.
\newblock The risks of invariant risk minimization.
\newblock In \emph{International Conference on Learning Representations}, 2021.
\newblock URL \url{https://openreview.net/forum?id=BbNIbVPJ-42}.

\bibitem[Sagawa* et~al.(2020)Sagawa*, Koh*, Hashimoto, and Liang]{gdro}
Shiori Sagawa*, Pang~Wei Koh*, Tatsunori~B. Hashimoto, and Percy Liang.
\newblock Distributionally robust neural networks.
\newblock In \emph{International Conference on Learning Representations}, 2020.
\newblock URL \url{https://openreview.net/forum?id=ryxGuJrFvS}.

\bibitem[Sagawa et~al.(2020)Sagawa, Raghunathan, Koh, and Liang]{gauss1}
Shiori Sagawa, Aditi Raghunathan, Pang~Wei Koh, and Percy Liang.
\newblock An investigation of why overparameterization exacerbates spurious
  correlations.
\newblock In \emph{Proceedings of the 37th International Conference on Machine
  Learning}, ICML'20. JMLR.org, 2020.

\bibitem[Setlur et~al.(2023)Setlur, Dennis, Eysenbach, Raghunathan, Finn,
  Smith, and Levine]{setlur}
Amrith Setlur, Don Dennis, Benjamin Eysenbach, Aditi Raghunathan, Chelsea Finn,
  Virginia Smith, and Sergey Levine.
\newblock Bitrate-constrained dro: Beyond worst case robustness to unknown
  group shifts, 2023.

\bibitem[Shah et~al.(2020)Shah, Tamuly, Raghunathan, Jain, and
  Netrapalli]{pitfalls}
Harshay Shah, Kaustav Tamuly, Aditi Raghunathan, Prateek Jain, and Praneeth
  Netrapalli.
\newblock The pitfalls of simplicity bias in neural networks.
\newblock In \emph{Proceedings of the 34th International Conference on Neural
  Information Processing Systems}, NIPS'20, Red Hook, NY, USA, 2020. Curran
  Associates Inc.
\newblock ISBN 9781713829546.

\bibitem[Shalev-Shwartz and Ben-David(2014)]{nfl2}
Shai Shalev-Shwartz and Shai Ben-David.
\newblock \emph{Understanding Machine Learning: From Theory to Algorithms}.
\newblock Cambridge University Press, USA, 2014.
\newblock ISBN 1107057132.

\bibitem[Sohoni et~al.(2020)Sohoni, Dunnmon, Angus, Gu, and R\'{e}]{george}
Nimit Sohoni, Jared~A. Dunnmon, Geoffrey Angus, Albert Gu, and Christopher
  R\'{e}.
\newblock No subclass left behind: Fine-grained robustness in coarse-grained
  classification problems.
\newblock In \emph{Proceedings of the 34th International Conference on Neural
  Information Processing Systems}, NIPS'20, Red Hook, NY, USA, 2020. Curran
  Associates Inc.
\newblock ISBN 9781713829546.

\bibitem[Sohoni et~al.(2022)Sohoni, Sanjabi, Ballas, Grover, Nie, Firooz, and
  Ré]{barack}
Nimit~S. Sohoni, Maziar Sanjabi, Nicolas Ballas, Aditya Grover, Shaoliang Nie,
  Hamed Firooz, and Christopher Ré.
\newblock Barack: Partially supervised group robustness with guarantees, 2022.

\bibitem[Teney et~al.(2022)Teney, Abbasnejad, Lucey, and van~den
  Hengel]{evading-sb}
Damien Teney, Ehsan Abbasnejad, Simon Lucey, and Anton van~den Hengel.
\newblock Evading the simplicity bias: Training a diverse set of models
  discovers solutions with superior ood generalization.
\newblock In \emph{Proceedings of the IEEE/CVF Conference on Computer Vision
  and Pattern Recognition (CVPR)}, pages 16761--16772, June 2022.

\bibitem[Utama et~al.(2020)Utama, Moosavi, and Gurevych]{utama}
Prasetya~Ajie Utama, Nafise~Sadat Moosavi, and Iryna Gurevych.
\newblock Mind the trade-off: Debiasing {NLU} models without degrading the
  in-distribution performance.
\newblock In \emph{Proceedings of the 58th Annual Meeting of the Association
  for Computational Linguistics}, pages 8717--8729, Online, July 2020.
  Association for Computational Linguistics.
\newblock \doi{10.18653/v1/2020.acl-main.770}.
\newblock URL \url{https://aclanthology.org/2020.acl-main.770}.

\bibitem[Valle-Perez et~al.(2019)Valle-Perez, Camargo, and
  Louis]{valle2018deep}
Guillermo Valle-Perez, Chico~Q Camargo, and Ard~A Louis.
\newblock Deep learning generalizes because the parameter-function map is
  biased towards simple functions.
\newblock \emph{International Conference on Learning Representations}, 2019.

\bibitem[Wah et~al.(2011)Wah, Branson, Welinder, Perona, and Belongie]{cub}
C.~Wah, S.~Branson, P.~Welinder, P.~Perona, and S.~Belongie.
\newblock {The Caltech-UCSD Birds-200-2011 Dataset}.
\newblock Technical Report CNS-TR-2011-001, California Institute of Technology,
  2011.

\bibitem[Wang et~al.(2019)Wang, Zhao, Yatskar, Chang, and Ordonez]{wang-iccv}
Tianlu Wang, Jieyu Zhao, Mark Yatskar, Kai-Wei Chang, and Vicente Ordonez.
\newblock Balanced datasets are not enough: Estimating and mitigating gender
  bias in deep image representations.
\newblock In \emph{International Conference on Computer Vision (ICCV)}, October
  2019.

\bibitem[Williams et~al.(2018)Williams, Nangia, and Bowman]{multinli}
Adina Williams, Nikita Nangia, and Samuel Bowman.
\newblock A broad-coverage challenge corpus for sentence understanding through
  inference.
\newblock In \emph{Proceedings of the 2018 Conference of the North American
  Chapter of the Association for Computational Linguistics: Human Language
  Technologies, Volume 1 (Long Papers)}, pages 1112--1122. Association for
  Computational Linguistics, 2018.
\newblock URL \url{http://aclweb.org/anthology/N18-1101}.

\bibitem[Wolpert(1996)]{nfl1}
David~H. Wolpert.
\newblock The lack of a priori distinctions between learning algorithms.
\newblock \emph{Neural Computation}, 8:\penalty0 1341--1390, 1996.
\newblock URL \url{https://api.semanticscholar.org/CorpusID:207609360}.

\bibitem[Xiao et~al.(2020)Xiao, Engstrom, Ilyas, and Madry]{imagenet9}
Kai Xiao, Logan Engstrom, Andrew Ilyas, and Aleksander Madry.
\newblock Noise or signal: The role of image backgrounds in object recognition,
  2020.

\bibitem[Zhang et~al.(2017)Zhang, Bengio, Hardt, Recht, and
  Vinyals]{zhang2017understanding}
Chiyuan Zhang, Samy Bengio, Moritz Hardt, Benjamin Recht, and Oriol Vinyals.
\newblock Understanding deep learning requires rethinking generalization.
\newblock In \emph{International Conference on Learning Representations}, 2017.
\newblock URL \url{https://openreview.net/forum?id=Sy8gdB9xx}.

\bibitem[Zhang et~al.(2022)Zhang, Lopez-Paz, and Bottou]{bonsai}
Jianyu Zhang, David Lopez-Paz, and Leon Bottou.
\newblock Rich feature construction for the optimization-generalization
  dilemma.
\newblock In Kamalika Chaudhuri, Stefanie Jegelka, Le~Song, Csaba Szepesvari,
  Gang Niu, and Sivan Sabato, editors, \emph{Proceedings of the 39th
  International Conference on Machine Learning}, volume 162 of
  \emph{Proceedings of Machine Learning Research}, pages 26397--26411. PMLR,
  17--23 Jul 2022.
\newblock URL \url{https://proceedings.mlr.press/v162/zhang22u.html}.

\bibitem[Zhou et~al.(2017)Zhou, Lapedriza, Khosla, Oliva, and Torralba]{places}
Bolei Zhou, Agata Lapedriza, Aditya Khosla, Aude Oliva, and Antonio Torralba.
\newblock Places: A 10 million image database for scene recognition.
\newblock \emph{IEEE Transactions on Pattern Analysis and Machine
  Intelligence}, 2017.

\end{thebibliography}

\newpage
\appendix
\section{Proofs for Section \ref*{sec:th-res}}
In this section, we present the proofs for the theoretical results in Section \ref{sec:th-res}.
\subsection{Proof of Proposition \ref*{prop:erm_gaussian}}
We first restate Proposition ~\ref{prop:erm_gaussian}:
\renewcommand{\thepropo}{\ref*{prop:erm_gaussian}}
\begin{propo}
ERM($\calM$) satisfies $\tfrac{\wa}{\wb}\!=\!\tfrac{\ma}{\mb'}\tfrac{\sib'^2}{\sia^2}$. When  $\tfrac{\ma}{\mb'}\tfrac{\sib'^2}{\sia^2}\!<\!1$, ERM($\calM_s$)$=\!\left[0,\tfrac{\mb'}{\sib^2}\right]$ (upto scaling). 
\end{propo}
\begin{proof}
We have:
\begin{align*}
    \E\,(\wa\fta+\wb\ftb-y)^2
    &=\wa^2(\sia^2+\ma^2)+\wb^2(\sib^2+\mb^2)+1-2\wa\ma-2\wb\mb'+2\wa\wb\ma\mb'.
\end{align*}
Since $\argmin_{\w\in\calM }\,\E\,(\wa\fta+\wb\ftb-y)^2$ is a convex problem, to find the minimizer we set gradients with respect to $\wa$ and $\wb$ to be $0$. 
Subsequently, we solve the resulting set of equations. By taking the gradient and setting it to 0, we obtain the following set of equations:
\begin{align}    2\wa(\sia^2+\ma^2)-2\ma+2\wb\ma\mb'&=0\label{eq1},\\    2\wb(\sib^2+\mb^2)-2\mb'+2\wa\ma\mb'&=0\label{eq2}.
\end{align}
From (\ref{eq2}), $\wb=\mb'\frac{1-\wa\ma}{\sib^2
+\mb^2}$. Substituting this in (\ref{eq1}) and solving for $\wa$, we get:
\begin{align*}
    w_1=\tfrac{\ma}{\sia^{2}}\left(\tfrac{1}{\tfrac{\ma^{2}}{\sia^{2}}+\tfrac{\mb'^{2}}{\sib'^2}+1}\right),
\end{align*}
where $\sib'^2=\sib^{2}+\mb^2-\mb'^2$.
Using this, we get the expression for $\wb$:

\begin{align*}
 w_2=\tfrac{\mb'}{\sib'^{2}}\left(\tfrac{1}{\tfrac{\ma^{2}}{\sia^{2}}+\tfrac{\mb'^{2}}{\sib'^{2}}+1}\right).
\end{align*}
Thus, for ERM($\calM$), $\tfrac{\wa}{\wb}=\tfrac{\ma}{\mb'}\tfrac{\sib'^2}{\sia^2}$.

Next, consider ERM over $\calM_s$. If $\wb=0$, we use (\ref{eq1}) and get $\left[\tfrac{\ma}{\sia^2+\ma^2},0\right]$ as the solution, for which the loss value is $\tfrac{\sia^2}{\sia^2+\ma^2}$. If $\wa=0$, we use (\ref{eq2}) and get $\left[0,\tfrac{\mb'}{\sib^2+\mb^2}\right]$ as the solution, for which the loss is $\tfrac{\sib'^2}{\sib'^2+\mb'^2}$. When $\tfrac{\ma^2}{\mb'^2}\tfrac{\sib'^2}{\sia^2}<1$, the latter has a smaller loss and thus, ERM($\calM_s$) is $\left[0,\tfrac{\mb'}{\sib^2}\tfrac{1}{1+\tfrac{\mb^2}{\sib^2}}\right]$.
\end{proof}
\subsection{Proof of Theorem 1}
\renewcommand{\thetheorem}{\ref*{thm:gauss1}}
\begin{theorem}
Let data be generated as per Assumption \ref{assumpt:gauss}. 
For $\nu=0.5\log(1+\eps^2)$ for some $\eps$:

1. When $\tfrac{\ma}{\mb'}\tfrac{\sib'^2}{\sia\sib}>\tfrac{1}{\eps}$, the solution to \eqref{cmi-gauss} 
is the same as ERM($\calM$), so $\tfrac{\wa}{\wb}=\tfrac{\ma}{\mb'}\tfrac{\sib'^2}{\sia^2}$.

2. Otherwise, $\wa$ is upweighted and the solution to \eqref{cmi-gauss} 
satisfies $\tfrac{|\wa|}{|\wb|}=\tfrac{1}{\eps}\tfrac{\sib}{\sia}$. 

\end{theorem}
\renewcommand{\thetheorem}{\arabic{theorem}}
\begin{proof} Consider the constraint $I(\wa\fta+\wb\ftb;\wb^*\ftb|y)\leq \nu$. As we are working with continuous random variables in this setup, we employ differential entropy for our entropy computations. The entropy of a Gaussian random variable $X$ with variance $\sigma^2$ is 
given by $H(X)=0.5(\log(2\pi\sigma^2)+1)$. Using this and the definitions of CMI and conditional entropy, we have:
\begin{align*}    I(\wa\fta+\wb\ftb,\wb^*\ftb|y)&=H(\wa\fta+\wb\ftb|y) - H(\wa\fta+\wb\ftb|y,\wb^*\ftb)\\
    &=H(\wa\fta+\wb\ftb|y) - H(\wa\fta|y)=\tfrac{1}{2}\log\left(\tfrac{\wa^2\sia^2+\wb^2\sib^2}{\wa^2\sia^2}\right).
\end{align*}
Using $\nu=0.5\log(1+\eps^2)$, the constraint becomes: $\tfrac{\wb^2\sib^2}{\wa^2\sia^2}\leq \eps^2$. Thus, (\ref{cmi-gauss}) reduces to solving:
\begin{align*}
\min_{\wa,\wb}&\E\,(\wa\fta+\wb\ftb-y)^2 \text{ s.t. } \tfrac{|\wb|\sib}{|\wa|\sia}\leq \eps. 
\end{align*}
If $\tfrac{\mb'}{\ma}\tfrac{\sia\sib}{\sib'^2}<\eps$, ERM($\calM$) satisfies the constraint and serves as the solution to (\ref{cmi-gauss}). Otherwise, since this is a convex optimization problem with an affine constraint, the constraint must be tight. Therefore, we determine the solution by finding ERM($\calM$) subject to $\tfrac{|\wb|\sia}{|\wa|\sib}=\eps$. The solution is given by:
\begin{align*}
    w_1=\tfrac{\mu_{1}}{\sigma_{1}^{2}}\left(\tfrac{\left(1+\tfrac{\mu_{2}'\sigma_{1}}{\mu_{1}\sigma_{2}}\eps\right)}{\tfrac{\mu_{1}^{2}}{\sigma_{1}^{2}}+1+\eps^{2}\left(\tfrac{\mu_{2}^{2}}{\sigma_{2}^{2}}+1\right)+\tfrac{2\eps\mu_{1}\mu_{2}'}{\sigma_{1}\sigma_{2}}}\right),\quad
     w_2=c\tfrac{\sigma_{1}}{\sigma_{2}}\tfrac{\mu_{1}}{\sigma_{1}^{2}}\left(\tfrac{\left(1+\tfrac{\mu_{2}'\sigma_{1}}{\mu_{1}\sigma_{2}}\eps\right)}{\tfrac{\mu_{1}^{2}}{\sigma_{1}^{2}}+1+\eps^{2}\left(\tfrac{\mu_{2}^{2}}{\sigma_{2}^{2}}+1\right)+\tfrac{2\eps\mu_{1}\mu_{2}'}{\sigma_{1}\sigma_{2}}}\right).
\end{align*}
\end{proof}
\vspace{-2mm}
\subsection{Proof of Theorem 2}
\renewcommand{\thetheorem}{\ref*{gauss-th-res2}}
\begin{theorem}
Let the data be generated as per Assumption \ref{assumpt:gauss} and \eqref{eq:x3}. Let $\nu\!=\!0.5\log(1+\eps^2)$ for some sufficiently small $\eps$, and $\tfrac{\sib^2}{\sib'^2}=\tfrac{\sic^2}{\sic'^2}$. Then the solution to \eqref{cmi-gauss2} satisfies: $
\tfrac{|\wb\mb'+\wc\mc'|}{|\wa|\sia}= \eps'$, where $\eps'=2\eps\sqrt{\tfrac{\mb'^2}{\sib^2}+\tfrac{\mc'^2}{\sic^2}}$. Moreover, 
$\wb,\wc\propto\eps'\wa$.
\end{theorem}
\renewcommand{\thetheorem}{\arabic{theorem}}
\begin{proof} Let $Z_1=\wa\fta+\wb\ftb+\wc\ftc$ and $Z_2=\wb^*\ftb+\wc^*\ftc$. Then,
\begin{align*}   
I(Z_1,\,Z_2|y)=H(Z_1|y)+H(Z_2|y)-H(Z_1,Z_2|y).
\end{align*}  
 Since all features are Gaussian, we have:
 \begin{align*}   H(Z_1|y)&=0.5(\log(\wa^2\sia^2+\wb^2\sib^2+\wc^2\sic^2)+\log(2\pi)+1),\\  H(Z_2|y)&=0.5(\log((\wb^*)^2\sib^2+(\wc^*)^2\sic^2)+\log(2\pi)+1),\\
     H(Z_1,Z_2|y)&=0.5 \log|K|+\log(2\pi)+1,
 \end{align*}
where $K$ is the covariance matrix of $Z_1$ and $Z_2$ conditioned on $y$. We can calculate $K$ as follows. Since all features are conditionally independent, we have: \begin{itemize}[leftmargin=*]
    \item $\E(Z_1-\E(Z_1))^2=\wa^2\sia^2+\wb^2\sib^2+\wc^2\sic^2$.
    \item $\E(Z_1-\E(Z_1))(Z_2-\E(Z_2))=\wb\wb^*\sib^2+\wc\wc^*\sic^2$.
    \item $\E(Z_2-\E(Z_2))^2=(\wb^*)^2\sib^2+(\wc^*)^2\sic^2$.    
\end{itemize}
Thus, $|K|=(\wa^2\sia^2+\wb^2\sib^2+\wc^2\sic^2)((\wb^*)^2\sib^2+(\wc^*)^2\sic^2)-(\wb\wb^*\sib^2+\wc\wc^*\sic^2)^2$. 
Using these, the constraint becomes:
\begin{align*}    \log\left(\tfrac{(\wa^2\sia^2+\wb^2\sib^2+\wc^2\sic^2)((\wb^*)^2\sib^2+(\wc^*)^2\sic^2)}{(\wa^2\sia^2+\wb^2\sib^2+\wc^2\sic^2)((\wb^*)^2\sib^2+(\wc^*)^2\sic^2)-(\wb\wb^*\sib^2+\wc\wc^*\sic^2)^2}\right)&\leq \log(1+\eps^2)\\
\implies \tfrac{(\wb\wb^*\sib^2+\wc\wc^*\sic^2)^2}{\wa^2\sia^2((\wb^*)^2\sib^2+(\wc^*)^2\sic^2)+\sib^2\sic^2(\wb^*\wc-\wc^*\wb)^2}&\leq \eps^2\\
\implies
\tfrac{(\wb\wb^*\sib^2+\wc\wc^*\sic^2)^2-c^2\sib^2\sic^2(\wb^*\wc-\wc^*\wb)^2}{\wa^2\sia^2((\wb^*)^2\sib^2+(\wc^*)^2\sic^2)}&\leq \eps^2.
\end{align*}  
Assume that $c$ is sufficiently small, \textit{i.e.}, $c\leq\tfrac{\sqrt{3}|\wb\wb^*\sib^2+\wc\wc^*\sic^2|}{2\sib\sic|\wb\wc^*-\wc\wb^*|}$. Then, we get the condition:
\begin{align*}
\tfrac{(\wb\wb^*\sib^2+\wc\wc^*\sic^2)^2}{\wa^2\sia^2((\wb^*)^2\sib^2+(\wc^*)^2\sic^2)}&\leq 4\eps^2\\
\implies \tfrac{\left|\wb\mb'\tfrac{\wb^*\sib^2}{\mb'}+\wc\mc'\tfrac{\wc^*\sic^2}{\mc'}\right|}{|\wa|\sia}&\leq 2\eps\sqrt{(\wb^*)^2\sib^2+(\wc^*)^2\sic^2}.
\end{align*}

When $[\wb^*,\wc^*]=v\left[{\mb'}/{\sib'^2},{\mc'}/{\sic'^2}\right]$, where $v$ is some constant, we get:
\begin{align*}
    \tfrac{\left|\wb\mb'\tfrac{\sib^2}{\sib'^2}+\wc\mc'\tfrac{\sic^2}{\sic'^2}\right|}{|\wa|\sia}&\leq 2\eps\sqrt{\tfrac{\mb'^2\sib^2}{\sib'^4}+\tfrac{\mc'^2\sic^2}{\sic'^4}}.
\end{align*}
Since $\tfrac{\sib^2}{\sib'^2}=\tfrac{\sic^2}{\sic'^2}$, we get:
\begin{align*}
    \tfrac{\left|\wb\mb'+\wc\mc'\right|}{|\wa|\sia}&\leq 2\eps\sqrt{\tfrac{\mb'^2}{\sib^2}+\tfrac{\mc'^2}{\sic^2}}=\eps'.
\end{align*}

When $\tfrac{\tfrac{\mb'^2}{\sib'^2}+\tfrac{\mc'^2}{\sic'^2}}{\tfrac{\ma}{\sia}}\leq c'$, ERM($\calM$), \textit{i.e.}, $[\wa,\wb,\wc]\propto \left[\tfrac{\ma}{\sia^2},\tfrac{\mb'}{\sib'^2},\tfrac{\mc'}{\sic'^2}\right]$ satisfies the constraint, and thus is the solution to (\ref{cmi-gauss2}). Otherwise, since (\ref{cmi-gauss2}) is a convex problem with an affine constraint, the constraint must be tight. Therefore, we find ERM($\calM$) subject to the equality constraint $|\wb\mb'+\wc\mc'|=\eps'\sia|\wa|$. The solution is given by:
\begin{align*}
    \wa=\tfrac{\ma}{\sia^2}\tfrac{1+\eps'\tfrac{\sia}{\ma}}{1+\tfrac{\ma^2}{\sia^2}+2\eps'\tfrac{\ma}{\sia}+(\eps')^2\tfrac{1-\tfrac{\mb'^2}{\sib^2+\mb^2}\tfrac{\mc'^2}{\sic^2+\mc^2}}{\tfrac{\mb'^2}{\sib^2+\mb^2}+\tfrac{\mc'^2}{\sic^2+\mc^2}-2\tfrac{\mb'^2}{\sib^2+\mb^2}\tfrac{\mc'^2}{\sic^2+\mc^2}}},
\end{align*}
\begin{align*}
\wb=\eps'\sia\wa\tfrac{\tfrac{\mb'}{\sib^2+\mb^2}\left(1-\tfrac{\mc'^2}{\sic^2+\mc^2}\right)}{\tfrac{\mb'^2}{\sib^2+\mb^2}+\tfrac{\mc'^2}{\sic^2+\mc^2}-2\tfrac{\mb'^2}{\sib^2+\mb^2}\tfrac{\mc'^2}{\sic^2+\mc^2}}, \quad \wc=\eps'\sia\wa\tfrac{\tfrac{\mc'}{\sic^2+\mc^2}\left(1-\tfrac{\mb'^2}{\sib^2+\mb^2}\right)}{\tfrac{\mb'^2}{\sib^2+\mb^2}+\tfrac{\mc'^2}{\sic^2+\mc^2}-2\tfrac{\mb'^2}{\sib^2+\mb^2}\tfrac{\mc'^2}{\sic^2+\mc^2}}.
\end{align*}
\end{proof}
\vspace{-2mm}
\subsection{Proof of Proposition 2}
\renewcommand{\thepropo}{\ref*{th:ood}}
\begin{propo}
Let $\text{ERM}(\mathcal{\mdl}_s)=M_s^*$. 
    Under Assumptions \ref{simple-complex-fts} and \ref{ucm-asm}, the solution to the problem:
    \begin{align}
    \tag{\ref{ood-th-eqn}}
        \argmin_{\mdl\in\mathcal{\mdl}}\E \, \ell_\mdl(Z) \text{ s.t. } I(\mdl;M_s^*|Y)=0
    \end{align}
    is $\mdl=\Phi^*$, the maximal invariant predictor.
\end{propo}
\renewcommand{\thepropo}{\arabic{propo}}
\begin{proof}
Using Assumption \ref{simple-complex-fts}, the class of simple models only contains variant predictors, so $\text{ERM}(\mathcal{\mdl}_s)=\Psi_s$. Consequently, the constraint in (\ref{ood-th-eqn}) can be written as $I(\mdl;\Psi_s|Y)=0$. 

Considering the set of candidate predictors for $M$, namely $\{0,\Phi_c,\Psi_s,\Psi_c\}$, we examine the CMI constraint for each. Using the definition of mutual information, we have $I(0,\Psi_s|Y)=0$ and $I(\Psi_s,\Psi_s|Y)=H(\Psi_s|Y)$. According to the definition of \textit{variant} predictor, $H(\Psi_s|Y)>H(\Psi_s|Y,E)\geq 0$.

From Assumption \ref{ucm-asm}, which states that the invariant and variant predictors are conditionally independent, we can deduce that $I(\Phi_c,\Psi_s|Y)=0$. From Assumption \ref{ucm-asm}, we also have $I(\Psi_c,\Psi_s|Y)>I(\Psi_c,\Psi_s|Y,E)\geq 0$. 
 
 Using these results, the feasible set is $[0,\Phi_c]$, which corresponds to the invariance set $\calI_E(\mathcal{\mdl})$. Consequently, problem (\ref{ood-th-eqn}) is equivalent to finding $\argmax_{\Phi\in \calI_E(\mathcal{\mdl})} I(Y ; \Phi)$. The solution to this problem is $\Phi_c$, which represents the MIP $\Phi^*$.
\end{proof}
\vspace{-2mm}
\section{Experimental Settings}
We begin by describing some common details and notation that we use throughout this section. As in the main text, we use $\lambda$ to represent the regularization strength for CMI. To ensure effective regularization, we adopt an epoch-dependent approach by scaling the regularization strength using the parameter $S$. Specifically, we set $\lambda=\lm_c\left(1+t/\sca\right)$ at epoch $t$. The temperature parameter $T$ is set as $12.5$ throughout the experiments. Additionally, we use $\lr$ to denote the learning rate, $\bs$ to denote the batch size, and $\lm_2$ to denote the weight decay parameter, which represents the strength of $\ell_2$-regularization. When using the Adam optimizer, we employ the default values for momentum.

The experiments on Slab data, CMNIST and CPMNIST data and Adult-Confounded data were implemented on Google Colab. The ImageNet-9 experiments were run on an AWS G4dn instance with one NVIDIA T4 GPU. For experiments on the subgroup robustness datasets and the Camelyon17-WILDS data, we used two NVIDIA V100 GPUs with 32 GB memory each. We only used CPU cores for the Bios data experiments.
\vspace{-2mm}
\subsection{Mitigating Simplicity Bias Experiments}
This section includes the details for the experiments showing that CMID mitigates simplicity bias where we use the Slab data and the ImageNet-9 data.
\vspace{-2mm}
\subsubsection{Slab Data}
\paragraph{Dataset.} All the features in the 3-Slab and 5-Slab data are in the range $[-1,1]$. The features are generated by defining the range of the slabs along each direction and then sampling points in that range uniformly at random. The base code for data generation came from the official implementation of \citep{pitfalls} available at \url{https://github.com/harshays/simplicitybiaspitfalls}. We consider $10^5$ training samples and $5\times10^4$ test samples. In both the cases, the linear margin is set as $0.05$. The 3-Slab data is 10-dimensional, where the remaining 8 coordinates are standard Gaussians, and are not predictive of the label. The slab margin is set as 0.075. The 5-Slab data is only 2-dimensional, and the slab margin is set as 0.14. 
\paragraph{Training.} We consider a linear model for the simple model and following \citep{pitfalls}, a 1-hidden layer NN with 100 hidden units as the final model (for both ERM and CMID). Throughout, we use SGD with $\bs=500,\lm_2=5\times10^{-4}$ for training. The linear model is trained with $\lr=0.05$, while the NN is trained with $\lr=0.005$. 

For the 3-Slab data, the models are trained for 300 epochs. We consider $\lm_c\in\{100,150,200\}$ and choose $\lm_c=150$ for the final result. For the 5-Slab data, the models are trained for 200 epochs and we use a 0.99 momentum in this case. We consider $\lm_c\in\{1000,2000,2500,3000\}$ and choose $\lm_c=3000$ for the final result. Note that we consider significantly high values of $\lm_c$ for this dataset compared to the rest because the simple model is perfectly predictive of the label in this case. This implies that its CMI with the final model is very small, and the regularization strength needs to be large in order for this term to contribute to the loss.

\vspace{-2mm}
\subsubsection{Texture vs Shape Bias on ImageNet-9}
\label{in_app}

We use ResNet50 pretrained on ImageNet data as the simple model, and train it on ImageNet-9 using Adam with $\lr=0.001, \bs=32, \lm_2=10^{-4}$ for $10$ epochs. We do not consider a simpler architecture and training from scratch since this pre-trained model already exhibits texture bias. 
For the final model, we consider the same model and parameters, except we use SGD with 0.9 momentum as the optimizer and $\lm_2=0.001$ for both ERM and \name and train for $10$ epochs. Values of \name specific parameters were $\lm_c=15,\sca=10$. For tuning, we consider $\lr\in\{10^{-5},10^{-4},10^{-3}\}$ for both the models and $\bs\in\{16,32\},\lm_2\in\{0.0001,0.001,0.01\}$ and $\lm_c\in\{0.5, 15, 25, 50\}$. We implement JTT to obtain the results. The first-stage model is trained for 1 epoch, and we consider the upsampling parameter \citep{jtt} $\lm_{up} \in \{5, 20, 50, 100\}$ for tuning. The other hyperparameters are the same as our approach.

\vspace{-2mm}
\subsection{Better OOD Generalization Experiments}
This section includes the details for the experiments showing that \name leads to better OOD generalization. For this, we used CMNIST and CPMNIST, Camelyon17-WILDS and Adult-Confounded datasets.

\vspace{-2mm}
\subsubsection{CMNIST and CPMNIST}
\label{cmnist-hp}
\paragraph{Dataset.} Following \citep{blood}, we use $25,000$ MNIST images (from the official train split) for each of the training environments, and the remaining $10,000$ images to construct a validation set. For both the test sets, we use the $10,000$ images from the official test split.
\paragraph{Training.}
The details about model architecture and parameters for training the simple model with ERM and the final model with CMID, for both the datasets, are listed in Table \ref{cmnist-params}. Following \citep{blood}, the MLP has one hidden layer with $390$ units and ReLU activation function. In both the cases, the simple model is trained for 4 epochs, while the final model is trained for 20 epochs. We choose the model with the smallest accuracy gap between the training and validation sets. For tuning, we consider the following values for each parameter: for the simple model, $\lr\in\{0.005,0.01,0.05\}$, $\lm_2\in\{0.001,0.005,0.01\}$, and for the final model, $\lr\in\{0.001,0.005\}$, $\lm_c\in[3,8]$ and $\sca\in[3,6]$, where lower values of $S$ were tried for higher values of $\lm_c$ and vice-versa. 

For the final results, we report the mean and standard deviation by averaging over 4 runs. For comparison, we consider the results reported by \citep{blood} for all methods, except EIIL \citep{eiil} and JTT \citep{jtt}. Results for EIIL are obtained by using their publicly available implementation for CMNIST data (available at \url{https://github.com/ecreager/eiil}), and incorporating the CPMNIST data into their implementation. The hyperparameter values in their implementation are kept the same. We implement JTT to obtain the results. We consider $\lr\in\{0.001,0.005,0.01\}$ and the parameter for upweighting minority groups \citep{jtt} $\lm_{up}\in\{5,10,15,20,25\}$ for tuning.
\begin{table}[h!]
\centering
\resizebox{0.85\columnwidth}{!}{%
\begin{tabular}{cccccccccccc}
\hline 
Dataset & Simple Model & Optimizer & $\lr$ & BS & $\lm_2$ & Final Model & Optimizer & $\lr$ & BS & $\lm_c$ & $\sca$ \\
\hline 
CMNIST & Linear & SGD & 0.01 & 64 & 0.005 & MLP & SGD & 0.001 & 64 & 4 & 4 \\
CPMNIST & Linear & SGD & 0.01 & 64 & 0.001 & MLP & SGD & 0.005 & 64 & 5 & 3 \\
\hline
\end{tabular}%
}
\caption{Training details for CMNIST and CPMNIST.}
\label{cmnist-params}
\end{table}
\vspace{-3mm}
\subsubsection{Camelyon17-WILDS}
\label{cam_app}
\paragraph{Dataset.} Camelyon17-WILDS \citep{wilds} contains $96\times96$ image patches which may or may not display tumor tissue in the central region. We use the same dataset as Bae et al. \citep{blood}, which includes $302,436$ training patches, $34,904$ OOD validation patches, and $85,054$ OOD test patches, where no two data-splits contain images from overlapping hospitals. We use the WILDS package, available at \url{https://github.com/p-lambda/wilds} for dataloading.
\paragraph{Training.} For the simple model, we train a \ts{2DConvNet1} model (see Section \ref{simple-models} for details) for 10 epochs. We use the Adam optimizer with $\lr=10^{-4},\bs=32,\lm_2 = 10^{-4}$. For the final model, we train a DenseNet121 (randomly initialized, no pretraining) for 5 epochs using SGD with 0.9 momentum with $\lr=10^{-4},\bs=32,\lm_2=0.01$ and $\lm_c=0.5,\sca=10$. We use the same $\bs$ and $\lm_2$ values as \citep{blood} for consistency. For tuning, we consider $\lr\in\{10^{-5},10^{-4},10^{-3}\}$ for both the models and $\lm_c\in\{0.5, 2, 5, 15\}$ for CMID. While \citep{blood} and \citep{wilds} use learning rates $10^{-5}$ and $0.001$, respectively, we found a learning rate of $10^{-4}$ was most suited for our approach. We select the model with the highest average accuracy on the validation set to report the final results, and report the mean and standard deviation by averaging over 3 runs. For comparison, we use the results reported by \citep{blood} for all the methods, except JTT. For the results for JTT, we implement the method and tune $\lm_{up}\in \{20, 50, 100\}$ and number of epochs to train the first-stage model as $\{1,2\}$. The rest of the hyperparameters are kept the same as our approach. We noticed that JTT \citep{jtt} had high variance in test accuracy across multiple runs for different random weight initialization. To account for this, we ran JTT and CMID over the same 8 seeds and randomly choose 3 of them to report the average accuracies over. We leave explorations on random weight initialization for JTT and similar approaches to future work. 

\subsubsection{Adult-Confounded}
\paragraph{Dataset.} The UCI Adult dataset \citep{uci1,uci2}, comprises $48, 842$ census records collected from the USA in 1994. Following \citep{eiil}, we use the original train/test splits from UCI Adult as well as the same subgroup sizes, but individual examples are under/over-sampled using importance weights based on the correlation on the original data and the desired correlation for the Adult-Confounded dataset. 
\paragraph{Training.} We use a linear model (with a bias term) as the simple model, and following \citep{eiil} use an Adagrad optimizer throughout. We use $\bs=50$. The simple model is trained for $50$ epochs, with $\lr=0.05, \lm_2=0.001$. Following \citep{eiil,lahoti}, we use a two-hidden-layer MLP architecture for the final model, with $64$ and $32$ hidden units, respectively. It is trained with $\lr=0.04, \lm_c=4, \sca=4$ for $10$ epochs. We also construct a small validation set from the train split by randomly selecting a small fraction of samples $(5-50)$ from each subgroup (depending on its size) and then upsampling these samples to get balanced subgroups of size $50$. We choose the model with lowest accuracy gap between the train and validation sets. For tuning, we consider $\lr\in\{0.01,0.02,0.03,0.04,0.05\}$ and $\lm_c,\sca\in\{3,4,5\}$. For the final results, we report the mean and accuracy by averaging over 4 runs. For comparison, we reproduced the results for ERM from \citep{eiil}, and thus, consider the values reported in \citep{eiil} for ERM, ARL and EIIL. We implement JTT to obtain the results. We use $LR=0.05$ and $\lambda_2=0.001$ to train the final model and tune the parameter for upweighting minority groups \citep{jtt} $\lm_{up}$ over $\{10,20,25,30\}$. The remaining parameters are kept the same as for our approach.  
\subsection{Subgroup Robustness Experiments}
\label{spurious_app}
This section includes the details for the experiments showing that CMID enhances subgroup robustness, where we use four benchmark datasets: Waterbirds, CelebA, MultiNLI and CivilComments.

\subsubsection{Model Architectures}
\label{simple-models}
In this section, we discuss the architectures we consider for the simple models for this task. For the two image datasets, a shallow 2D CNN is a natural choice for the simple model as 2D CNNs can capture local patterns and spatial dependencies in grid-like data. On the other hand, for the two text datasets with tokenized representations, we consider a shallow MLP or 1D CNN for the simple model. MLPs can capture high-level relationships between tokens by treating each token as a separate feature, while 1D CNNs can capture local patterns and dependencies in sequential data.

\begin{figure}[h!]
  \centering
  \begin{minipage}{0.55\columnwidth}
  \begin{subfigure}{\textwidth}
    \centering
    \includegraphics[height=38pt]{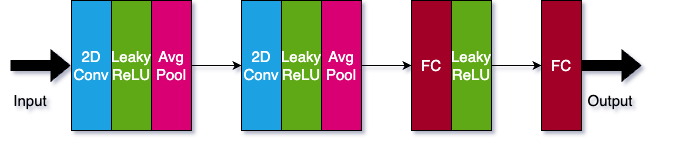}  
  \end{subfigure}
  \end{minipage}\hfill
  \begin{minipage}{0.4\columnwidth}
    \centering
 \begin{subfigure}{\textwidth}
    \centering
    \includegraphics[height=38pt]{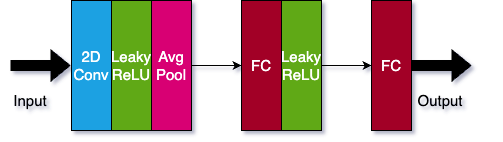}
  \end{subfigure}
  \end{minipage}
  \caption{Left: \ts{2DConvNet1} and Right: \ts{2DConvNet2} architectures.}
  \label{ConvNet}
\end{figure}
\vspace{-2mm}
\begin{figure}[h!]
  \centering
  \begin{minipage}{0.4\columnwidth}
  \begin{subfigure}{\textwidth}
    \centering
    \includegraphics[height=38pt]{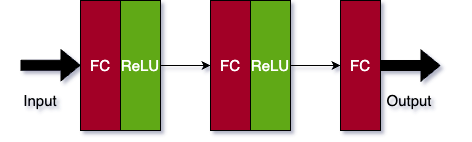}  
  \end{subfigure}
  \end{minipage}\hfill
  \begin{minipage}{0.55\columnwidth}
    \centering
 \begin{subfigure}{\textwidth}
    \centering
    \includegraphics[height=38pt]{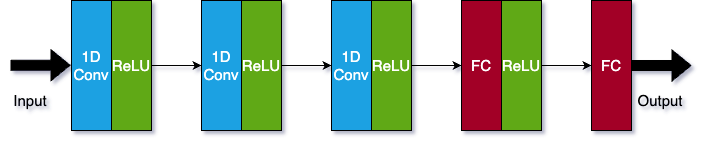}
  \end{subfigure}
  \end{minipage}
  \caption{Left: \ts{2MLP} and Right: \ts{1DConvNet} architectures.}
  \label{FCN}
\end{figure}

Next, we describe the details for the model architectures. Let $F$ denote the filter size and $C$ denote the number of output channels (for convolutional layers) or the output dimension (for linear/fully connected (FC) layers). Throughout, we use $F\!=\!2$ for the average pooling layers. Fig. \ref{ConvNet} shows the \ts{2DConvNet1} and the \ts{2DConvNet2} architecture, which were used as simple models for Waterbirds and CelebA, respectively. These were the only two architectures we considered for the 2DCNN on these datasets. In \ts{2DConvNet1}, the 2D convolutional layers use $F\!=\!7,C\!=\!10$ and $F\!=\!4,C\!=\!20$, respectively, while $C\!=\!2000$ for the FC layer. In the \ts{2DConvNet2} architecture, $F\!=\!5,C\!=\!10$ for the 2D convolutional layer and $C\!=\!500$ for the FC layer. Fig. \ref{FCN} shows the \ts{2MLP} and the \ts{1DConvNet} architecture, which were used as simple models for MultiNLI and CivilComments-WILDS, respectively. For tuning, we considered these models as well as a 1DCNN with one less 1D convolutional layer than \ts{1DConvNet} for both the datasets. In \ts{2MLP}, the FC layers use $C\!=\!100$ and $C\!=\!25$, respectively. In the \ts{1DConvNet} architecture, the 1D convolutional layers use $F\!=\!7,C\!=\!10$, $F\!=\!5,C\!=\!32$ and $F\!=\!5,C\!=\!64$, respectively, while $C\!=\!500$ for the FC layer.

\subsubsection{Training Details}
We utilize the official implementation of \citep{gdro} available at \url{https://github.com/kohpangwei/group_DRO} as baseline code and integrate our approach into it. Most hyperparameter values are kept unchanged, and we list the the important parameters along with model architectures for all the datasets in Table \ref{subgroup-exps}. For the simple models, we consider shallow 2D CNNs for the image datasets, and MLP and 1D CNN for the text data, as discussed in the previous section. In all cases, the simple model is trained for 20 epochs. For the final model, following \citep{gdro}, we use the Pytorch \tsc{torchvision} implementation of ResNet50 \citep{resnet50} with pretrained weights on ImageNet data for the image datasets, and the Hugging Face \tsc{pytorch-transformers} implementation of the BERT \tsc{bert-base-uncased} model, with pretrained weights \citep{devlin-bert} for the language-based datasets.

\begin{table}[h!]
\centering
\resizebox{\columnwidth}{!}{%
\begin{tabular}{cccccccccccccc}
\hline 
Dataset & Simple Model & Optimizer & $\lr$ & BS & $\lm_2$ & Final Model & Optimizer & $\lr$ & BS & $\lm_2$ & $\lm_c$ & $\sca$ & $\#$ epochs\\
\hline 
Waterbirds & \ts{2DConvNet1} & Adam & $10^{-5}$ & $32$ & $10^{-4}$ & ResNet50 & SGD & $5\!\times\! 10^{-4}$ & $128$ & $10^{-4}$ & $20$ & $4$ & $100$\\
CelebA & \ts{2DConvNet2} & Adam & $10^{-5}$ & $32$ & $5\!\times\! 10^{-4}$ & ResNet50 & SGD & $3\!\times\! 10^{-4}$ & $128$ & $0.001$ & $10$ & $5$ & $50$\\
MultiNLI & \ts{2MLP} & Adam & $0.005$ & $16$ & $10^{-4}$ & BERT & AdamW & $5\!\times\!10^{-5}$ & $32$ & $0$ & $75$ & $10$ & $5$ \\
CivilComments & \ts{1DConvNet} & Adam & $10^{-4}$ & $16$ & $10^{-4}$ & BERT & AdamW & $10^{-5}$ & $32$ & $0.001$ & $25$ & $10$ & $10$\\
\hline
\end{tabular}%
}
\caption{Training details for subgroup robustness datasets.}
\label{subgroup-exps}
\end{table}
\begin{wraptable}[6]{r}{0.64\columnwidth}
    \centering
    \vspace{-3mm}
    \resizebox{0.64\columnwidth}{!}{%
\begin{tabular}{cccc}
\hline Dataset & $\lr$ & $\lm_c$ & $\sca$\\
\hline Waterbirds, CelebA & $[1,5]\times10^{-4}$ & $\{10,15,20,25,50,75\}$ & $\{4,5,6,8,10\}$ \\
MultiNLI, CivilComments & $\{1,2,5\}\times10^{-5}$ & $\{10,25,50,75\}$ & $\{10\}$ \\
\hline
\end{tabular}
}\caption{Values considered for tuning the hyperparameters for training the final model for the four subgroup robustness datasets.}
\label{sg-tuning}
\end{wraptable}
Table \ref{sg-tuning} shows the values of $\lr,\lm_c$ and $\sca$ we consider for tuning for the final model. Following \citep{gdro}, we keep $\lm_2=0$ for MultiNLI. For the rest, we consider $\lm_2\in\{0.0001,0.0005,0.001\}$. For results, we choose the model with the best worst-group accuracy on the validation set. For comparison, we consider the values reported in \citep{jtt}.
\subsection{Bias in Occupation Prediction Experiment}
We use a version of the Bios data shared by the authors of \citep{bios-data}. We used the official implementation of \citep{snob}, available at \url{https://github.com/pinkvelvet9/snobpaper}, to obtain results and for comparison purposes. In this implementation, they consider 25 occupations and train separate one-vs-all linear classifiers for each occupation based on word embeddings to make predictions. We directly used their implementation to obtain results for ERM and Decoupled \citep{decoupled} on the data. 

For our approach, we employed linear models for both the simple model and the final model. We directly regularized the CMI with respect to the ERM from their implementation. The final model was trained for $5$ epochs using SGD with $\lr=0.1,\bs=128,\lm_c=5,\sca=5$. We only tuned the $\lr$ for this case, considering values of $0.05$ and $0.1$.

\subsection{Effect of Simple Model Architecture Experiment}
\begin{wraptable}[5]{r}{0.55\columnwidth}
\vspace{-3mm}
\centering
\resizebox{0.55\columnwidth}{!}{%
\begin{tabular}{lccccccc}    
\hline 
 & \multicolumn{4}{c}{ Waterbirds } & \multicolumn{2}{c}{ CMNIST} \\
 &   Linear & Shallow CNN   & ResNet18 & ResNet50 & Linear & MLP  \\
\hline 
LR  & $5\times10^{-5}$ & $10^{-5}$& $5\times10^{-6}$& $ 2\times10^{-5}$& $0.01 $ & $0.01 $\\
\hline
\end{tabular}}
\caption{Training details for the simple models.}
    \label{sma-tuning}
\end{wraptable}
We tune the learning rate (LR) for training the simple model (as listed in Table \ref{sma-tuning}). The rest of the hyperparameter values for training the final model are kept consistent across each dataset (as listed in Table \ref{cmnist-params} for CMNIST and Table \ref{sg-tuning} for Waterbirds).

\end{document}